\documentclass[11pt]{article}

\usepackage[final]{acl}
\usepackage{times}
\usepackage{latexsym}
\usepackage[T1]{fontenc}
\usepackage[utf8]{inputenc}
\usepackage{microtype}
\usepackage{inconsolata}

\usepackage{array}
\usepackage{booktabs}
\usepackage{multirow}
\usepackage{threeparttable}
\usepackage{graphicx}
\usepackage{xcolor}
\usepackage{amsmath}
\usepackage{amssymb}
\usepackage{enumitem}
\usepackage{subcaption}
\usepackage{tikz}
\usetikzlibrary{positioning, calc}
\usepackage{pgfplots}
\usepackage{algorithm}
\usepackage{algorithmic}
\usepackage{CJKutf8}
\usepackage{placeins}
\usepackage{dblfloatfix}

\definecolor{aclblue}{HTML}{000099}
\hypersetup{
    colorlinks=true,
    linkcolor=aclblue,
    citecolor=aclblue,
    urlcolor=aclblue
}

\newcommand{\tgcot}{\textsc{PG-CoT}}
\newcommand{\krl}{\textsc{K-RL}}

\title{Syndrome, Synergy, and Safety: Structured Reasoning and Knowledge-Driven Alignment for TCM Prescription Generation}

\begin{document}
\author{  Zheng Chen\textsuperscript{1} \quad ZhiCheng Du\textsuperscript{1} \quad Haoxuan Li\textsuperscript{1} \quad Peiwu Qin\textsuperscript{2,\dag} \thanks{\textsuperscript{2,\dag}Corresponding author: pwqin1979@gmail.com.}\\
  \textsuperscript{1}Tsinghua University \quad \textsuperscript{2}Guangdong Provincial Laboratory of Traditional Chinese Medicine Hengqin \\
  }

\maketitle

\begin{abstract}
Applying large language models to Traditional Chinese Medicine (TCM) prescription generation reveals three clinically critical gaps: models produce end-to-end mappings without auditable reasoning following the \emph{li-fa-fang-yao} paradigm (SR Gap), treat each encounter in isolation without follow-up adjustment via \emph{sui zheng jia jian} (LA Gap), and fail to enforce absolute contraindication rules such as \emph{Shi Ba Fan} (SC Gap). We propose a progressive four-stage framework (SFT $\to$ PG-CoT $\to$ Dynamic $\to$ K-RL) that addresses each gap: PG-CoT constrains CoT distillation under the \emph{li-fa-fang-yao} paradigm to produce auditable diagnostic chains, Dynamic SFT models patient trajectories with explicit transition reasoning, and K-RL encodes deterministic pharmacological rules as rule-based DPO preference signals. Across 12 fine-tuned models and 6 zero-shot baselines, our framework substantially improves prescription quality over zero-shot baselines—with a 7B model (Mistral-7B) surpassing zero-shot GPT-5 on all three TCM evaluation metrics.

\end{abstract}

\section{Introduction}
\label{sec:intro}

\begin{figure*}[t]
    \centering
    \includegraphics[width=\linewidth]{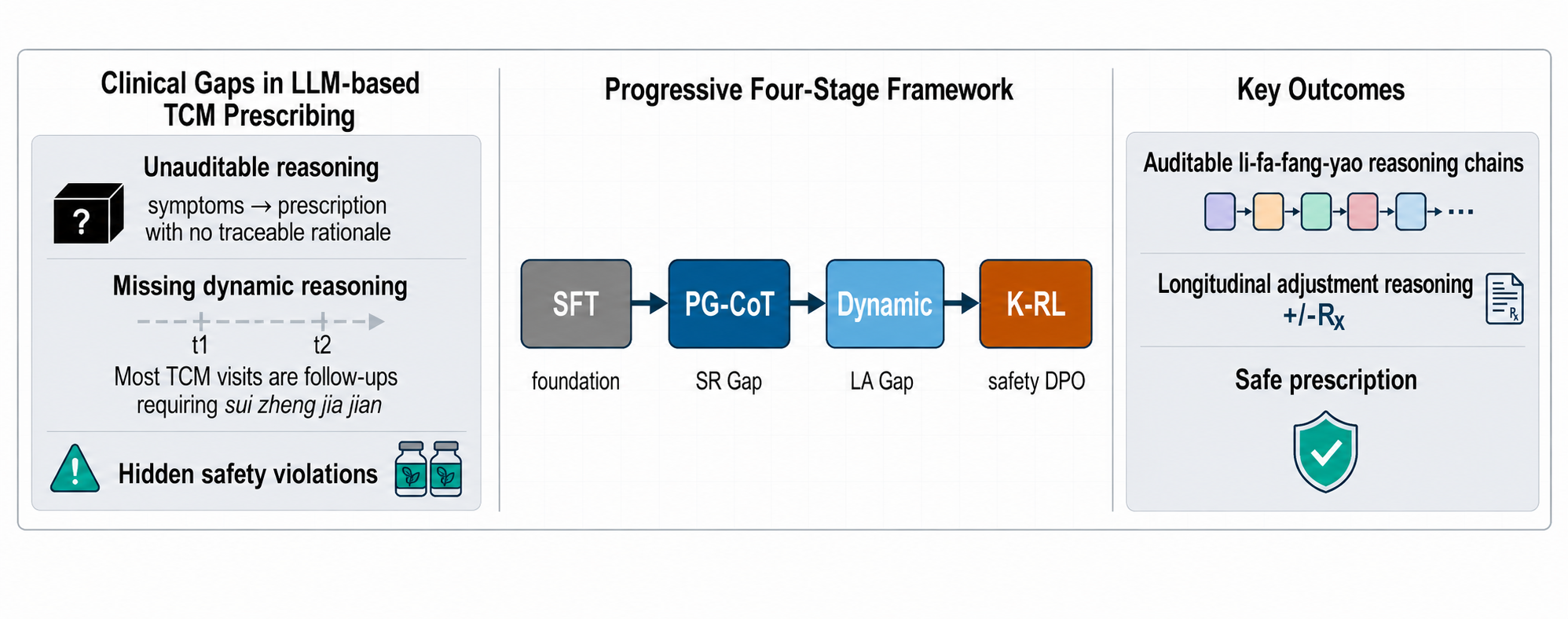}
    \caption{Visual abstract. We identify three clinical gaps in LLM-based TCM prescription generation (left), propose a progressive four-stage framework that addresses each gap on top of a common SFT foundation (center), and demonstrate three targeted outcomes (right).}
    \label{fig:teaser}
\end{figure*}

Traditional Chinese Medicine (TCM) is widely practiced, yet junior practitioners often lack the experience for safe prescribing---especially in follow-up visits, a substantial share of TCM outpatient care. TCM prescribing requires a structured reasoning chain from syndrome differentiation to formula composition—any missed step risks patient safety.

Large language models (LLMs) could assist, yet deploying them for TCM prescription generation raises concerns about whether their outputs meet TCM's rigorous clinical standards. Current medical LLMs exhibit three clinically critical gaps (detailed clinical background in Appendix~\ref{app:clinical_background}):

\paragraph{Structured Reasoning Gap (SR Gap).}
LLMs produce end-to-end symptom-to-prescription mappings without any mechanism to audit whether the output follows the \emph{li-fa-fang-yao} chain, the stepwise reasoning path from pathomechanism to medication that is treated as the unified logic linking syndrome differentiation to prescribing~\citep{jiang2012syndrome,xie2023formulas}.

\paragraph{Longitudinal Adaptation Gap (LA Gap).}
Follow-up visits require \emph{sui zheng jia jian} (adjusting herbs and dosages as symptoms evolve). Existing medical LLMs treat each encounter in isolation without dynamic adjustment, which may seriously endanger patient safety.
\vspace{-0.5em}
\paragraph{Safety Compliance Gap (SC Gap).}
Medical LLMs do not strictly enforce TCM's absolute \emph{Shi Ba Fan} (Eighteen Incompatibilities) and \emph{Shi Jiu Wei} (Nineteen Mutual Antagonisms) rules~\citep{long2013incompatibility}, which may yield hidden safety violations---life-threatening contraindicated combinations.

To address these three gaps, we propose a four-stage framework---baseline supervised fine-tuning (SFT) $\to$ Paradigm-Guided Chain-of-Thought (\tgcot{}) $\to$ Dynamic $\to$ Knowledge-Driven Reinforcement Learning (\krl{})---that progressively builds interpretability, longitudinal adaptation, and hard safety on top of a common foundation (Figure~\ref{fig:teaser}). Our contributions are:
\begin{enumerate}[nosep,leftmargin=*]
    \item \textbf{\tgcot{}} that combines CoT distillation with the \emph{li-fa-fang-yao} paradigm to produce auditable diagnostic chains, improving reasoning transparency.
    \item \textbf{Dynamic follow-up modeling} that incorporates \emph{sui zheng jia jian} transition reasoning on patient trajectories, enabling longitudinal prescription adjustment.
    \item \textbf{\krl{}} that encodes TCM pharmacological rules (\emph{Shi Ba Fan}, \emph{Shi Jiu Wei}) as rule-based DPO preference signals constructed from synthetic safety pairs, reducing contraindication violations.
    \item A comprehensive evaluation across \textbf{12 fine-tuned models} and \textbf{6 zero-shot baselines} with stage-wise ablation and adversarial probing, demonstrating that paradigm-constrained fine-tuning outperforms zero-shot high-parameter models.
\end{enumerate}

\begin{figure*}[t]
    \centering
    \includegraphics[width=\linewidth]{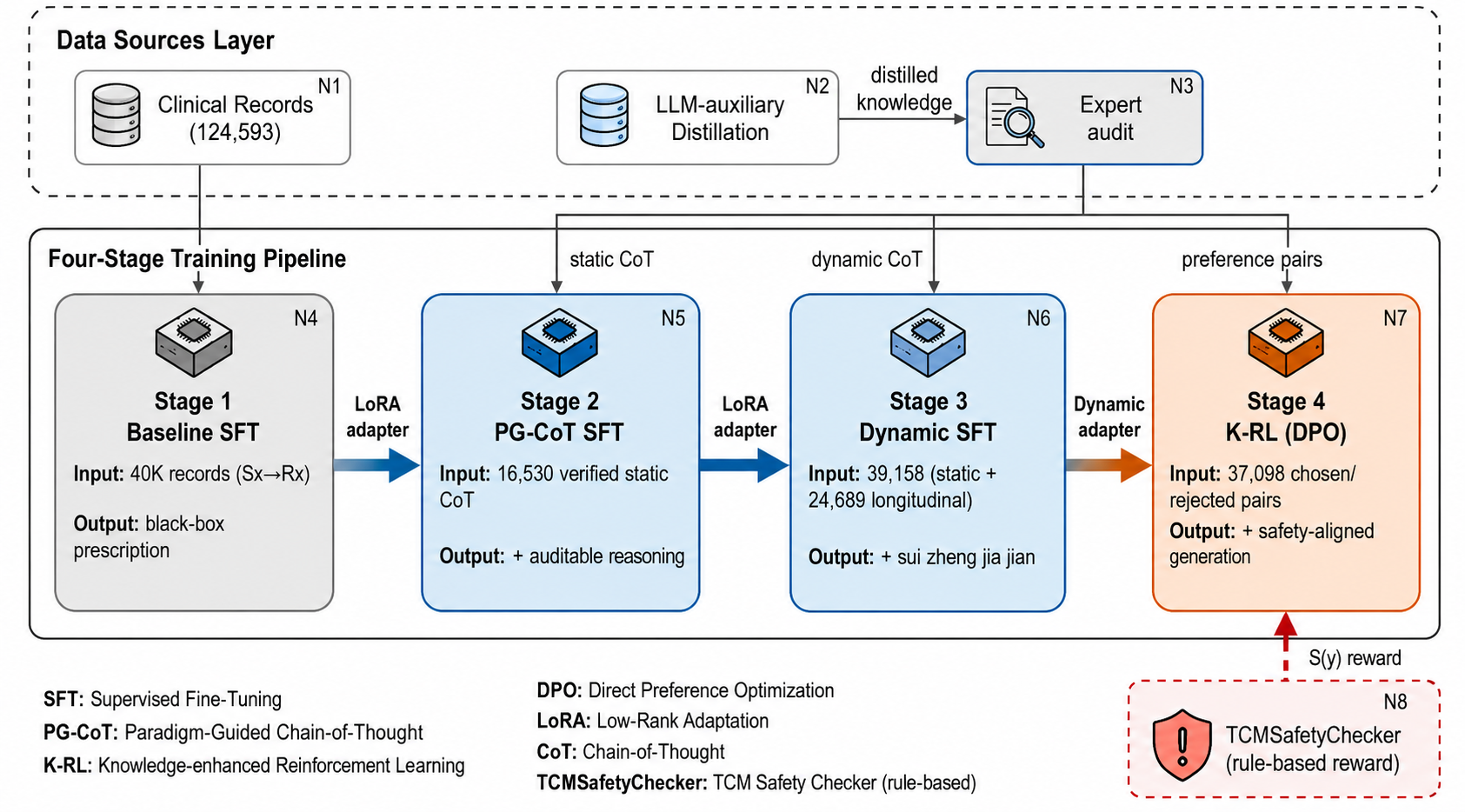}
    \caption{Four-stage training pipeline. Baseline SFT establishes basic prescription ability from 40K clinical records; \tgcot{} injects auditable reasoning via 16.5K \emph{li-fa-fang-yao} CoT samples; Dynamic SFT extends to longitudinal trajectories; \krl{} applies DPO with synthetic preference pairs whose reward is the rule-based safety score $S(\mathbf{y})$ from TCMSafetyChecker. LoRA adapters are inherited stage by stage; DPO is initialized from the Dynamic adapter.}
    \label{fig:framework}
\end{figure*}

Figure~\ref{fig:framework} overviews the full four-stage pipeline.

\section{Related Work}
\label{sec:related}

\subsection{Medical NLP and TCM}

LLMs have shown strong performance in medical question answering~\citep{singhal2023large} and diagnostic reasoning~\citep{tu2024towards}, with English systems like Med-PaLM~2~\citep{singhal2025medpalm2} reaching expert levels. Chinese models such as HuatuoGPT~\citep{wang2023huatuogpt} focus on general medical QA. TCM-specific NLP remains sparse: prior work addresses NER, relation extraction, and syndrome classification~\citep{zhang2022tcmie}, while recent TCM-centric LLMs such as BenTsao~\citep{wang2023bentsao} fine-tune on domain knowledge but do not tackle prescription generation or structured reasoning. \citet{yue2024tcmbench} propose a TCM benchmark focused on knowledge recall rather than prescription quality. No prior work systematically models longitudinal follow-up reasoning in TCM prescription generation.

\subsection{Chain-of-Thought and Knowledge Distillation}

Chain-of-thought prompting~\citep{wei2022chain} and distillation~\citep{ho2023large,hsieh2023distilling} improve performance on complex tasks via intermediate reasoning. In medicine, CoT has been used for English QA~\citep{singhal2023large,wang2023huatuogpt}, but the reasoning is free-form. For TCM, free-form CoT risks generating clinically illogical rationales because the reasoning must follow the fixed \emph{li-fa-fang-yao} paradigm. Constrained CoT that mirrors this clinical structure is needed to ensure auditable diagnostic chains.

\subsection{Safety Alignment and RLAIF}

RLHF~\citep{ouyang2022training} and DPO~\citep{rafailov2023direct} align LLMs with human preferences. RLAIF~\citep{bai2022constitutional,lee2024rlaif} replaces human annotators with LLM judges, inheriting their biases. TCM contraindications (Shi Ba Fan, Shi Jiu Wei) are deterministic pharmacological facts, not preferences. Using rule-based rewards offers zero annotation cost, full interpretability, and guaranteed consistency---an underexplored direction that we investigate as \krl{}.

\subsection{Longitudinal Clinical Modeling}

Longitudinal patient modeling has been explored in medical dialogue corpora~\citep{zeng2020meddialog} and conversational diagnostic AI~\citep{tu2025conversational}. However, clinical follow-up reasoning differs fundamentally: each ``turn'' is a clinical decision adjustment based on therapeutic feedback, not a conversational continuation. Existing models do not support explicit symptom-evolution comparison and justification of herb/dosage changes, which is exactly the \emph{sui zheng jia jian} capability we address.

\section{Method}
\label{sec:method}

\subsection{Task Definition}
\label{sec:task}

We formulate the \textbf{longitudinal TCM prescription generation} task as follows. Given a patient record $\mathbf{x}$ consisting of symptoms, tongue and pulse descriptions, and optional prior prescription history, the model must produce:
\begin{enumerate}[nosep,leftmargin=*]
    \item A \textbf{reasoning chain} $\mathbf{r} = (r_1, r_2, \ldots, r_K)$ following the \emph{li-fa-fang-yao} diagnostic paradigm (syndrome differentiation $\to$ treatment principle $\to$ formula rationale $\to$ safety verification),
    \item A \textbf{prescription} $\mathbf{y} = \{(h_i, d_i)\}_{i=1}^{N}$ of herb--dosage pairs.
\end{enumerate}

In the follow-up setting, the model additionally receives the initial prescription $\mathbf{y}^{(0)}$ and must reason about symptom evolution before generating an adjusted prescription $\mathbf{y}^{(1)}$. We factor the joint probability as:
\begin{equation}
    P(\mathbf{r}, \mathbf{y} \mid \mathbf{x}) = P(\mathbf{r} \mid \mathbf{x}) \cdot P(\mathbf{y} \mid \mathbf{r}, \mathbf{x})
\end{equation}
where $\mathbf{r}$ conditions prescription generation on explicit diagnostic reasoning, providing intermediate supervision.

Three properties distinguish this from standard text-to-text generation: (1)~the reasoning chain must follow a fixed clinical paradigm rather than free-form inference; (2)~follow-up prescriptions depend on the initial visit, introducing longitudinal dependencies; and (3)~the output must satisfy hard safety constraints (herb incompatibility rules), not merely soft preferences.

\subsection{PG-CoT: Paradigm-Guided Chain-of-Thought Distillation}
\label{sec:tgcot}

A naive application of chain-of-thought to TCM would encourage free-form reasoning. However, clinical validity in TCM requires adherence to the \emph{li-fa-fang-yao} paradigm: any deviation from this structured chain can produce plausible-sounding but clinically unjustifiable rationales. To quantify this, we conducted a pilot experiment (Appendix~\ref{app:pgcot_pilot}) comparing free-form CoT against a variant that explicitly follows the \emph{li-fa-fang-yao} structure. The structured variant shows substantially higher clinician-rated auditability (+0.29) and lower logical inconsistency ($-$0.12) while also improving prescription quality (Appendix~\ref{app:pgcot_pilot}), confirming that the paradigm constraint acts as a performance-enhancing inductive bias rather than a limitation.

\begin{figure*}[t]
    \centering
    \includegraphics[width=\linewidth]{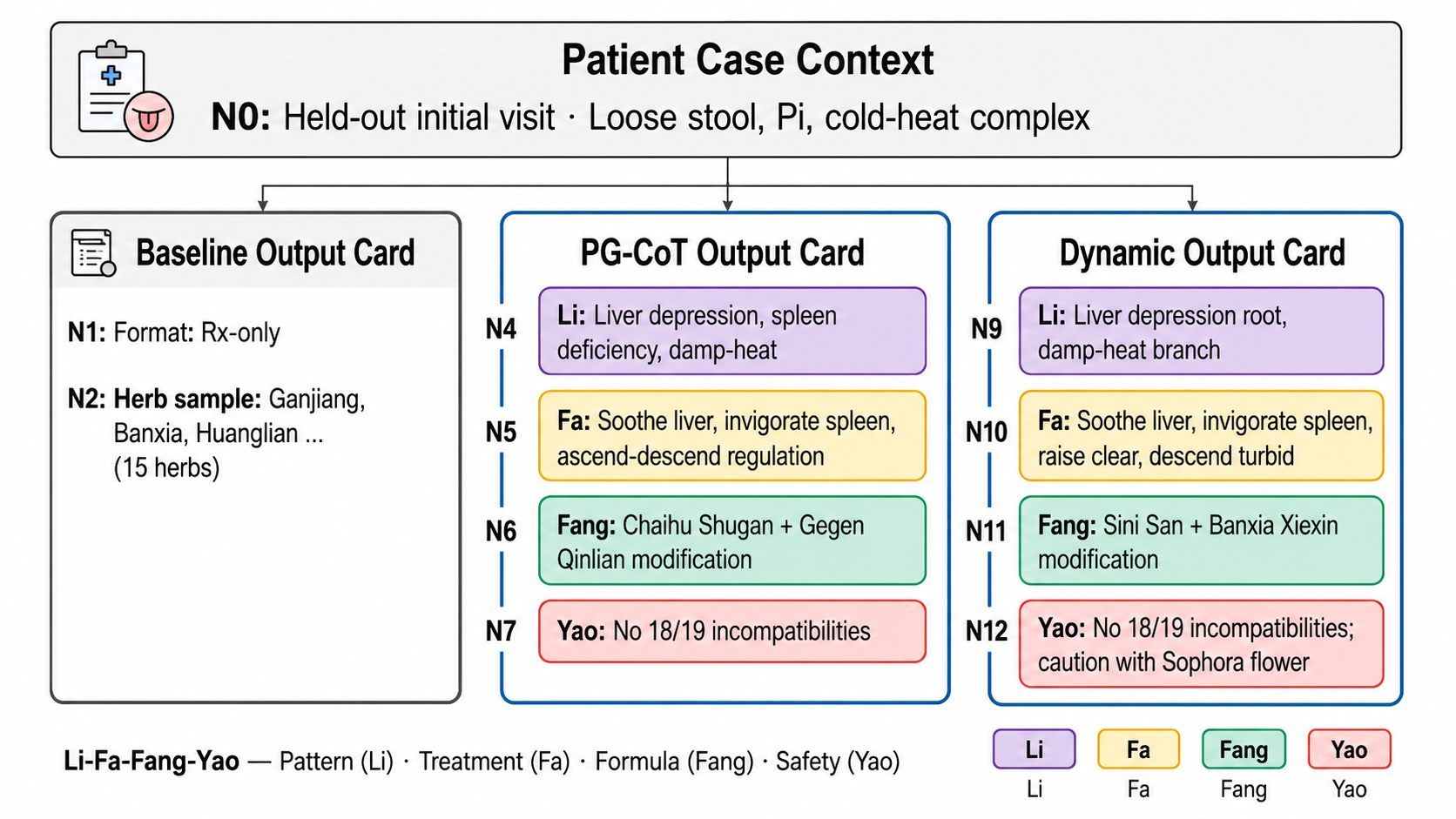}
    \caption{Output morphology comparison on one held-out initial visit (Mistral-7B). Baseline produces a prescription list only; PG-CoT and Dynamic SFT yield structured \emph{li-fa-fang-yao} chains with syndrome analysis, treatment principle, formula rationale, and safety verification.}
    \label{fig:stage_case}
\end{figure*}

Based on this evidence, we propose \textbf{PG-CoT} (Paradigm-Guided Chain-of-Thought). PG-CoT distills structured diagnostic chains from a strong reasoning model (DeepSeek-R1) into smaller student models, while strictly enforcing the \emph{li-fa-fang-yao} sequence. Unlike general CoT distillation~\citep{ho2023large,hsieh2023distilling} that encourages free-form reasoning, PG-CoT produces clinically auditable rationales where each step corresponds to a verifiable diagnostic decision (Figure~\ref{fig:stage_case}).

\paragraph{Static CoT structure.}
For initial visits, each training sample follows:
\[
    \mathbf{x}_{\text{sx}} \longrightarrow \underbrace{r_{\text{syn}} \to r_{\text{princ}} \to r_{\text{form}} \to r_{\text{safe}}}_{\text{structured reasoning chain}} \longrightarrow \mathbf{y}_{\text{rx}}
\]
where each $r_k$ is a structured text segment:
\begin{itemize}[nosep,leftmargin=*]
    \item \textbf{Syndrome analysis} ($r_{\text{syn}}$): integrates symptoms, tongue coating, and pulse qualities to identify the pathogenesis and derive the zheng (pattern), including zang-fu organ vacuity/repletion.
    \item \textbf{Treatment principle} ($r_{\text{princ}}$): establishes the therapeutic strategy (e.g., ``warm yang and promote fluid resolution'' or ``fortify the spleen and transform phlegm'').
    \item \textbf{Formula rationale} ($r_{\text{form}}$): explains the jun-chen-zuo-shi (king-minister-assistant-courier) compatibility logic of each core herb.
    \item \textbf{Safety verification} ($r_{\text{safe}}$): checks for contraindication rule violations (Section~\ref{sec:krl}).
\end{itemize}

\subsection{Dynamic Regime: Longitudinal Follow-up Modeling}
\label{sec:dynamic}

The defining clinical skill in follow-ups is \emph{sui zheng jia jian}: adjusting herbs and dosages based on symptom evolution. We address the LA Gap (Section~\ref{sec:intro}) through longitudinal follow-up modeling with dynamic CoT.

Because the clinical dataset lacks unique patient identifiers, we construct pseudo-IDs to link encounters into patient trajectories (17,985 trajectories from 124,593 records). We filter follow-up transitions by Jaccard similarity of herb sets ($0.3 \leq J < 0.9$, clinically meaningful adjustment), yielding 24,689 dynamic samples; full details are in Appendix~\ref{app:data_pipeline}.

\paragraph{Dynamic CoT structure.}
For follow-up visits, the reasoning chain is extended to reason about the transition:
\[
\begin{aligned}
(\mathbf{x}^{(1)}, \mathbf{y}^{(0)}) &\longrightarrow
\underbrace{r_{\text{evol}} \to r_{\text{adj}} \to r_{\text{new}} \to r_{\text{safe}}}_{\text{dynamic CoT chain}} \\
&\longrightarrow \mathbf{y}^{(1)}
\end{aligned}
\]
where $r_{\text{evolution}}$ compares symptoms between visits (improvement, persistence, new manifestations), $r_{\text{adjustment}}$ explains specific herb additions/removals (e.g., ``heat has resolved, therefore remove Shi Gao; spleen deficiency persists, therefore add Bai Zhu''), and $r_{\text{new\_formula}}$ summarizes the new prescription's compatibility logic.

The key distinction from static CoT is the introduction of \textbf{longitudinal dependency}: the model must understand not just \emph{what to prescribe} but \emph{why this prescription should differ from the previous one}.

\subsection{K-RL: Knowledge-Driven Preference Alignment}
\label{sec:krl}

TCM contraindications (e.g., \emph{Shi Ba Fan} pairs) are deterministic pharmacological facts rather than learned preferences. We address the SC Gap (Section~\ref{sec:intro}) by exploring \krl{} (Knowledge-Driven Reinforcement Learning) as an initial attempt to replace human preference annotation with rule-based domain knowledge as the reward signal; the resulting safety effects are analyzed in Section~\ref{sec:results}.

\paragraph{TCMSafetyChecker.}
We implement a rule-based safety scorer grounded in TCM pharmacological hard knowledge:
\begin{itemize}[nosep,leftmargin=*]
    \item \textbf{Shi Ba Fan} (Eighteen Incompatibilities): 6 core pairs with derived synonyms (e.g., licorice $\leftrightarrow$ Gan Sui / Da Ji / Yuan Hua / Hai Zao; Aconite $\leftrightarrow$ Bei Mu / Gua Lou / Ban Xia / Bai Ji). Penalty: $-0.5$.
    \item \textbf{Shi Jiu Wei} (Nineteen Mutual Antagonisms): 10 pairs (e.g., Ding Xiang $\leftrightarrow$ Yu Jin; Ren Shen $\leftrightarrow$ Wu Ling Zhi). Penalty: $-0.3$.
    \item \textbf{Toxic herb check}: herbs classified as ``highly toxic'' (e.g., raw Chuan Wu, Ma Qian Zi). Penalty: $-0.1$.
\end{itemize}
Synonym expansion prevents alias-based evasion (e.g., Hei Shun Pian $\to$ Fu Zi, Bei Xi Xin $\to$ Xi Xin). The final safety score is:
\begin{equation}
    S(\mathbf{y}) = \max\left(0, \, 1 + \sum_{v \in \mathcal{V}} w_v \cdot \mathbb{1}[v \in \mathbf{y}]\right)
\end{equation}
where $\mathcal{V}$ is the set of detected violations and $w_v$ is the penalty weight.

\paragraph{DPO preference pair construction.}
We construct preference pairs for Direct Preference Optimization~\citep{rafailov2023direct} as follows:
\begin{itemize}[nosep,leftmargin=*]
    \item \textbf{Chosen} ($\mathbf{y}_w$): LLM-assisted prescriptions that pass both the TCMSafetyChecker with $S(\mathbf{y}) = 1.0$ and human review (37,884 samples before decontamination; 37,098 after).
    \item \textbf{Rejected} ($\mathbf{y}_l$): the same prescriptions with synthetically injected contraindication herbs (e.g., adding Gan Sui to a prescription containing licorice).
    \item \textbf{Filtering}: 2,101 outputs that contained safety violations are excluded from the chosen set, ensuring that positive examples are unambiguously safe.
\end{itemize}

This synthetic-negative strategy avoids manual preference annotation and produces unambiguous contrast pairs, but its narrow distribution may not cover the adversarial scenarios encountered at evaluation time, a potential factor in K-RL's model-dependent effects (Section~\ref{sec:adversarial}; Limitations).

\paragraph{Training procedure.}
DPO is initialized from the \emph{Dynamic SFT} adapter (Section~\ref{sec:dynamic}) to preserve reasoning and longitudinal capabilities. The DPO objective is:
\begin{align}
    \mathcal{L}_{\text{DPO}} &= -\mathbb{E}\Big[\log \sigma\big(\beta \log \tfrac{\pi_\theta(\mathbf{y}_w | \mathbf{x})}{\pi_{\text{ref}}(\mathbf{y}_w | \mathbf{x})} \notag \\
    &\quad - \beta \log \tfrac{\pi_\theta(\mathbf{y}_l | \mathbf{x})}{\pi_{\text{ref}}(\mathbf{y}_l | \mathbf{x})}\big)\Big]
\end{align}
We set $\beta = 0.1$ and learning rate $5 \times 10^{-6}$, and train for only 1 epoch to avoid alignment tax (the degradation of model capabilities from overfitting on preference data).

\subsection{Training Overview}
\label{sec:training_overview}

All stages share a parameter-efficient fine-tuning backbone~\citep{hu2022lora}; GPU, quantization, hyperparameters, and scale-adaptive settings are in Appendix~\ref{app:hyperparams}.
\section{Experimental Setup}
\label{sec:setup}

We evaluate our four-stage pipeline on 12 fine-tuned models via QLoRA, targeting the three clinical gaps in Section~\ref{sec:intro} (SR, LA, and SC) through stage-wise ablation on a unified held-out test set of 871 cases. We additionally compare against 6 zero-shot baselines (Section~\ref{sec:sota_comparison}). Training details are in Appendix~\ref{app:hyperparams}, prompt templates in Appendix~\ref{app:prompts}, and reproducibility information in Appendix~\ref{app:reproducibility}.

\subsection{Data and Models}
\label{sec:data_models}

Our dataset is constructed from 124,593 TCM outpatient records, progressively filtered into four experimental datasets (Table~\ref{tab:dataset}; details in Appendix~\ref{app:data_pipeline}). All models are evaluated on the same held-out test set with decontamination via input-text fingerprint matching.

\begin{table}[t]
\centering
{\small
\begin{tabular}{lrr}
\toprule
Dataset & Train & Test \\
\midrule
Baseline (Sx $\to$ Rx) & 40,000 & 871 \\
PG-CoT (Sx $\to$ Rsn + Rx) & 16,530 & 871 \\
Dynamic (Sx + FU $\to$ Rsn + Rx) & 39,158 & 871 \\
K-RL / DPO (safety pairs) & 37,098 & 871 \\
\bottomrule
\end{tabular}
}
\caption{Dataset statistics. Sx = symptoms, FU = follow-up, Rsn = reasoning. All training counts are after train/test decontamination. Baseline is sampled to 40K from 94{,}986 decontaminated records. All stages share the same 871-case test set.}
\label{tab:dataset}
\end{table}

We evaluate 12 models spanning six families and three scale tiers: \textbf{Small} ($<$7B): Qwen3.5-0.8B, Qwen3.5-2B, Gemma-2-2B, Phi-3-Mini-3.8B; \textbf{Medium} (7--14B): Qwen3.5-9B, LLaMA-3.1-8B, Mistral-7B-v0.3, DeepSeek-R1-Distill-Llama-8B (DS-R1-Distill-8B), Gemma-2-9B; \textbf{Large} ($>$14B): Qwen3.5-27B, Gemma-2-27B, Mistral-Small-24B. The complete model list is in Appendix~\ref{app:models}; structured output compliance analysis is in Appendix~\ref{app:small_models}.

We additionally compare against 6 zero-shot baselines (Table~\ref{tab:sota_comparison}): 3 high-parameter general-purpose models (GPT-5, DeepSeek-V3, LLaMA-4-Scout) and 3 TCM-specific models (Baichuan2-7B, HuatuoGPT, BenTsao). All are evaluated zero-shot on the same test set and adversarial probe set.

\subsection{Evaluation Metrics}
\label{sec:metrics}

To simultaneously assess prescription accuracy and auditability under professional TCM standards, we adopt three metrics: PQS, CQS, and VR. Design rationale and implementation details are in Appendix~\ref{app:metric_rationale}.

\paragraph{PQS.}
\begin{equation}
\begin{split}
    \text{PQS} =\;& 0.30 F1_H + 0.35 \text{Acc}_{\text{dose}} + 0.15 J \\
    &+ 0.10 \text{OK}_{\text{cnt}} + 0.10 S
\end{split}
\end{equation}
where $F1_H$: herb-set F1; $\text{Acc}_{\text{dose}}$: dosage accuracy (40\% tolerance); $J$: herb-set Jaccard; $\text{OK}_{\text{cnt}}$: herb count reasonableness; $S$: safety score (Section~\ref{sec:krl}).

\paragraph{CQS.}
\begin{equation}
\begin{split}
    \text{CQS} =\;& 0.6 \cdot \text{PQS}_{\text{base}} + 0.4 \cdot A_{\text{reason}}, \\
    A_{\text{reason}} =\;& \sum_{k} w_k \cdot \cos(\mathbf{e}_{k}^{\text{gt}}, \mathbf{e}_{k}^{\text{pred}})
\end{split}
\end{equation}
where $\text{PQS}_{\text{base}}$ is the baseline SFT PQS (format-stable across stages), and $A_{\text{reason}}$ is the section-weighted cosine similarity between predicted and ground-truth CoT segments (section weights in Appendix~\ref{app:metric_rationale}). Embeddings: Qwen3-Embedding-8B. For non-CoT outputs, we report PQS only.

\paragraph{VR.}
Adversarial probe set $\mathcal{P}$: 53 prompts (24 direct, 17 contextual, 12 open-generation), validated with 100\% trigger rate.
\begin{equation}
    \text{VR} = \frac{1}{|\mathcal{P}|} \sum_{p \in \mathcal{P}} \mathbb{1}\!\left[S(\hat{\mathbf{y}}_p) < 1\right]
\end{equation}
Lower VR is better. Evaluated on 10 fine-tuned models across all four stages.

\paragraph{Format bias caveat.}
PQS comparisons between baseline and CoT models must account for a \textbf{format bias}: CoT outputs embed prescriptions in long structured text, causing parser mis-extraction that biases baseline PQS upward (Appendix~\ref{app:format_bias}).

\FloatBarrier

\begin{table*}[!t]
\centering
\setlength{\tabcolsep}{4pt}
\begin{threeparttable}
{\small
\begin{tabular*}{\linewidth}{@{\extracolsep{\fill}} l l c c | c c c @{}}
\toprule
 & & \multicolumn{2}{c}{\textbf{PQS $\uparrow$}} & \multicolumn{3}{c}{\textbf{CQS $\uparrow$}} \\
\cmidrule(lr){3-4} \cmidrule(lr){5-7}
Model & Scale & Zero-shot & Base & PG-CoT & Dyn & K-RL \\
\midrule
\multicolumn{7}{l}{\textit{Small models ($<$7B)}} \\
Qwen3.5-0.8B & 0.8B & 0.344 & \textbf{0.639} & 0.707 & \textbf{0.723} & 0.716 \\
Qwen3.5-2B   & 2B   & 0.472 & \textbf{0.699} & 0.711 & \textbf{0.715} & 0.702 \\
Gemma-2-2B   & 2B   & 0.160 & \textbf{0.604} & 0.387 & 0.376 & \textbf{0.553} \\
Phi-3-Mini   & 3.8B & 0.108 & \textbf{0.643} & 0.422 & 0.442 & \textbf{0.501} \\
\midrule
\multicolumn{7}{l}{\textit{Medium models (7--14B)}} \\
LLaMA-3.1-8B    & 8B  & 0.200 & \textbf{0.710} & 0.727 & \textbf{0.731} & 0.724 \\
Mistral-7B      & 7B  & 0.113 & \textbf{0.701} & 0.728 & 0.732 & \textbf{0.766} \\
DS-R1-Distill-8B & 8B  & 0.111 & \textbf{0.560} & 0.566 & 0.570 & \textbf{0.597} \\
Gemma-2-9B      & 9B  & 0.367 & \textbf{0.717} & 0.721 & 0.728 & \textbf{0.729} \\
Qwen3.5-9B      & 9B  & \textbf{0.514} & 0.449 & \textbf{0.584} & 0.522 & 0.519\tnote{1} \\
\midrule
\multicolumn{7}{l}{\textit{Large models ($>$14B)}} \\
Mistral-24B   & 24B & 0.396 & \textbf{0.721} & 0.723 & 0.718 & \textbf{0.731} \\
Qwen3.5-27B  & 27B & 0.433 & \textbf{0.566} & 0.572 & 0.577 & \textbf{0.581} \\
Gemma-2-27B  & 27B & 0.353 & \textbf{0.683} & 0.676 & 0.683 & \textbf{0.701} \\
\bottomrule
\end{tabular*}
}
\begin{tablenotes}
\small
\item All models are fine-tuned via QLoRA and evaluated on the same 871-case held-out test set with decontamination. ``Base'' = baseline SFT; ``Dyn'' = Dynamic SFT; ``K-RL'' = DPO on Dynamic adapter. Stage-wise PQS and $A_{\text{reason}}$ in Appendix~\ref{app:full_results}. PQS comparisons across stages must account for format bias (Appendix~\ref{app:format_bias}).
\item[1] Qwen3.5-9B shows zero-shot PQS exceeding baseline SFT PQS, attributed to strong Chinese-centric pretraining combined with verbose SFT output reducing extraction accuracy; see Appendix~\ref{app:scaling} for discussion.
\end{tablenotes}
\caption{Main results (PQS and CQS) on fine-tuned models across training stages. Zero-shot PQS provided for reference.}
\label{tab:main_results}
\end{threeparttable}
\end{table*}

\begin{table*}[!t]
\centering
\setlength{\tabcolsep}{5pt}
\begin{threeparttable}
{\small
\begin{tabular*}{\linewidth}{@{\extracolsep{\fill}} l c c c c @{}}
\toprule
 & \multicolumn{4}{c}{\textbf{VR $\downarrow$}} \\
\cmidrule(lr){2-5}
Model & Base & PG-CoT & Dyn & K-RL \\
\midrule
\multicolumn{5}{l}{\textit{Small ($<$7B)}} \\
Qwen3.5-0.8B & 0.385 & 0.297 & 0.372 & \textbf{0.283} \\
Gemma-2-2B & 0.189 & 0.190 & 0.194 & \textbf{0.186} \\
Phi-3-Mini & \textbf{0.382} & 0.401 & 0.410 & 0.386 \\
\midrule
\multicolumn{5}{l}{\textit{Medium (7--14B)}} \\
LLaMA-3.1-8B & 0.113 & 0.079 & 0.115 & \textbf{0.038} \\
Mistral-7B & 0.432 & 0.419 & 0.338 & \textbf{0.271} \\
DS-R1-Distill-8B & 0.394 & \textbf{0.376} & 0.394 & 0.381 \\
Gemma-2-9B & 0.340 & \textbf{0.094} & 0.132 & 0.189 \\
\midrule
\multicolumn{5}{l}{\textit{Large ($>$14B)}} \\
Qwen3.5-27B & 0.376 & 0.311 & 0.358 & \textbf{0.299} \\
Gemma-2-27B & 0.412 & 0.358 & 0.403 & \textbf{0.338} \\
Mistral-24B & 0.340 & 0.257 & 0.232 & \textbf{0.211} \\
\bottomrule
\end{tabular*}
}
\begin{tablenotes}
\small
\item VR is evaluated on 53 adversarial probes (24 direct, 17 contextual, 12 open-generation) with 100\% trigger rate. ``Base'' = baseline SFT; ``Dyn'' = Dynamic SFT; ``K-RL'' = DPO on Dynamic adapter. Qwen3.5-2B (small) and Qwen3.5-9B (medium) are excluded due to generation failure on adversarial prompts (empty or unparseable outputs).
\end{tablenotes}
\caption{Adversarial safety results (VR) on fine-tuned models across training stages.}
\label{tab:adversarial}
\end{threeparttable}
\end{table*}

\begin{table*}[t]
\centering
\setlength{\tabcolsep}{4pt}
\begin{threeparttable}
{\small
\begin{tabular*}{\linewidth}{@{\extracolsep{\fill}} l r c c c @{}}
\toprule
Model & Params & PQS $\uparrow$ & CQS $\uparrow$ & VR $\downarrow$ \\
\midrule
\multicolumn{5}{l}{\textit{General high-parameter models (zero-shot)}} \\
GPT-5~\citep{openai2025gpt5} & --- & 0.664 & 0.671 & 0.285 \\
DeepSeek-V3~\citep{deepseek2024v3} & 671B & 0.612 & 0.632 & 0.263 \\
LLaMA-4-Scout~\citep{meta2025llama4} & 109B & 0.573 & 0.571 & 0.332 \\
\midrule
\multicolumn{5}{l}{\textit{TCM-specific models (zero-shot)}} \\
Baichuan2-7B~\citep{yang2023baichuan2} & 7B & 0.213 & 0.310 & 0.334 \\
HuatuoGPT~\citep{wang2023huatuogpt} & 7B & 0.201 & 0.192 & 0.296 \\
BenTsao~\citep{wang2023bentsao} & 7B & 0.243 & 0.332 & 0.198 \\
\midrule
\multicolumn{5}{l}{\textit{Ours (fine-tuned, K-RL stage)}} \\
Gemma-2-2B & 2B & 0.604 & 0.553 & \textbf{0.186} \\
Mistral-7B & 7B & 0.701 & \textbf{0.766} & 0.271 \\
Mistral-24B & 24B & \textbf{0.721} & 0.731 & 0.211 \\
\bottomrule
\end{tabular*}
}
\begin{tablenotes}
\small
\item All baselines are evaluated zero-shot on the same 871-case test set and 53 adversarial probes. CQS for zero-shot baselines is computed from their free-form outputs using the same structured parsing pipeline; outputs that cannot be parsed into the \emph{li-fa-fang-yao} format receive reduced $A_{\text{reason}}$ scores. Our models use the K-RL (final) training stage. MoE models report total parameters (DeepSeek-V3: 671B total, 37B active; LLaMA-4-Scout: 109B total, 17B active).
\end{tablenotes}
\caption{Comparison results (PQS, CQS, and VR) with high-parameter models.}
\label{tab:sota_comparison}
\end{threeparttable}
\end{table*}

\section{Results and Analysis}
\label{sec:results}

\subsection{Main Results}
\label{sec:main_results}

Tables~\ref{tab:main_results}--\ref{tab:adversarial} present results across 12 fine-tuned models and four training stages; Table~\ref{tab:sota_comparison} compares representative fine-tuned models against zero-shot baselines. Notably, \textbf{Mistral-7B at only 7B parameters surpasses zero-shot GPT-5 on all three TCM evaluation metrics}, demonstrating that paradigm-constrained fine-tuning at a modest scale can exceed the zero-shot capability of far larger models.

\paragraph{SR Gap: PG-CoT delivers auditable, high-quality reasoning.}
Domain SFT is the prerequisite foundation; on top of it, PG-CoT produces structured \emph{li-fa-fang-yao} chains with high reasoning alignment (Figure~\ref{fig:stage_case}) and CQS that matches or exceeds baseline PQS despite format bias (Appendix~\ref{app:format_bias}). Our fine-tuned models substantially outperform both high-parameter general-purpose models and TCM-specific models on prescription quality and reasoning alignment (Table~\ref{tab:sota_comparison}).

\paragraph{LA Gap: Dynamic SFT and K-RL progressively improve longitudinal adaptation.}
Dynamic SFT preserves clinical quality while enabling \emph{sui zheng jia jian} reasoning, and K-RL further amplifies gains. Medium-scale models benefit most from the full pipeline, improving monotonically across all stages (Table~\ref{tab:ablation}).

\paragraph{SC Gap: Structured reasoning and K-RL substantially reduce safety violations.}
\label{sec:adversarial}
PG-CoT's dedicated $r_{\text{safety}}$ verification sharply reduces adversarial VR, and K-RL delivers the most consistent VR reductions on models with high baseline vulnerability. Our fine-tuned models achieve lower VR than all zero-shot baselines, including models orders of magnitude larger (Table~\ref{tab:sota_comparison}).

\subsection{Comparison with Zero-Shot Baselines}
\label{sec:sota_comparison}

Table~\ref{tab:sota_comparison} compares representative fine-tuned models against high-parameter general-purpose and TCM-specific models in a zero-shot setting (full results in Appendix~\ref{app:zeroshot}). Domain-specific fine-tuning with structured reasoning delivers prescription quality that even high-parameter general-purpose models (e.g., DeepSeek-V3 at 671B) cannot match zero-shot. TCM-specific models (7B scale), despite incorporating medical knowledge during training, lack the \emph{li-fa-fang-yao} paradigm constraint and longitudinal reasoning that our framework provides. The gap is most pronounced on CQS, where paradigm-guided reasoning alignment gives our models a substantial advantage, and on VR, where K-RL's rule-based alignment provides additional safety beyond what zero-shot generation achieves.

\subsection{Stage-by-Stage Ablation}
\label{sec:ablation}

\begin{table}[!t]
\centering
{\small
\begin{tabular}{lcccc}
\toprule
Stage & PQS & Herb F1 & CQS & Key \\
\midrule
\multicolumn{5}{l}{\textit{Mistral-7B}} \\
A$\to$B & $-$0.296 & $-$0.243 & ---$\to$0.728 & Interp. \\
B$\to$C & $-$0.014 & $-$0.013 & 0.728$\to$0.732 & Longit. \\
C$\to$D & $-$0.041 & $-$0.038 & 0.732$\to$0.766 & K-RL \\
\midrule
\multicolumn{5}{l}{\textit{Mistral-24B}} \\
A$\to$B & $-$0.341 & $-$0.293 & ---$\to$0.723 & Interp. \\
B$\to$C & $-$0.005 & $-$0.002 & 0.723$\to$0.718 & Longit. \\
C$\to$D & +0.018 & +0.022 & 0.718$\to$0.731 & K-RL \\
\midrule
\multicolumn{5}{l}{\textit{DS-R1-Distill-8B}} \\
A$\to$B & $-$0.292 & $-$0.101 & ---$\to$0.566 & Interp. \\
B$\to$C & +0.381 & +0.394 & 0.566$\to$0.570 & Longit. \\
C$\to$D & $-$0.044 & $-$0.068 & 0.570$\to$0.597 & K-RL \\
\bottomrule
\end{tabular}
}
\caption{Stage-by-stage ablation. A = Baseline, B = PG-CoT, C = Dynamic, D = K-RL. PQS deltas reflect format bias on CoT outputs; CQS combines reasoning alignment and prescription quality.}
\label{tab:ablation}
\end{table}

\begin{figure}[t]
    \centering
    \includegraphics[width=\linewidth]{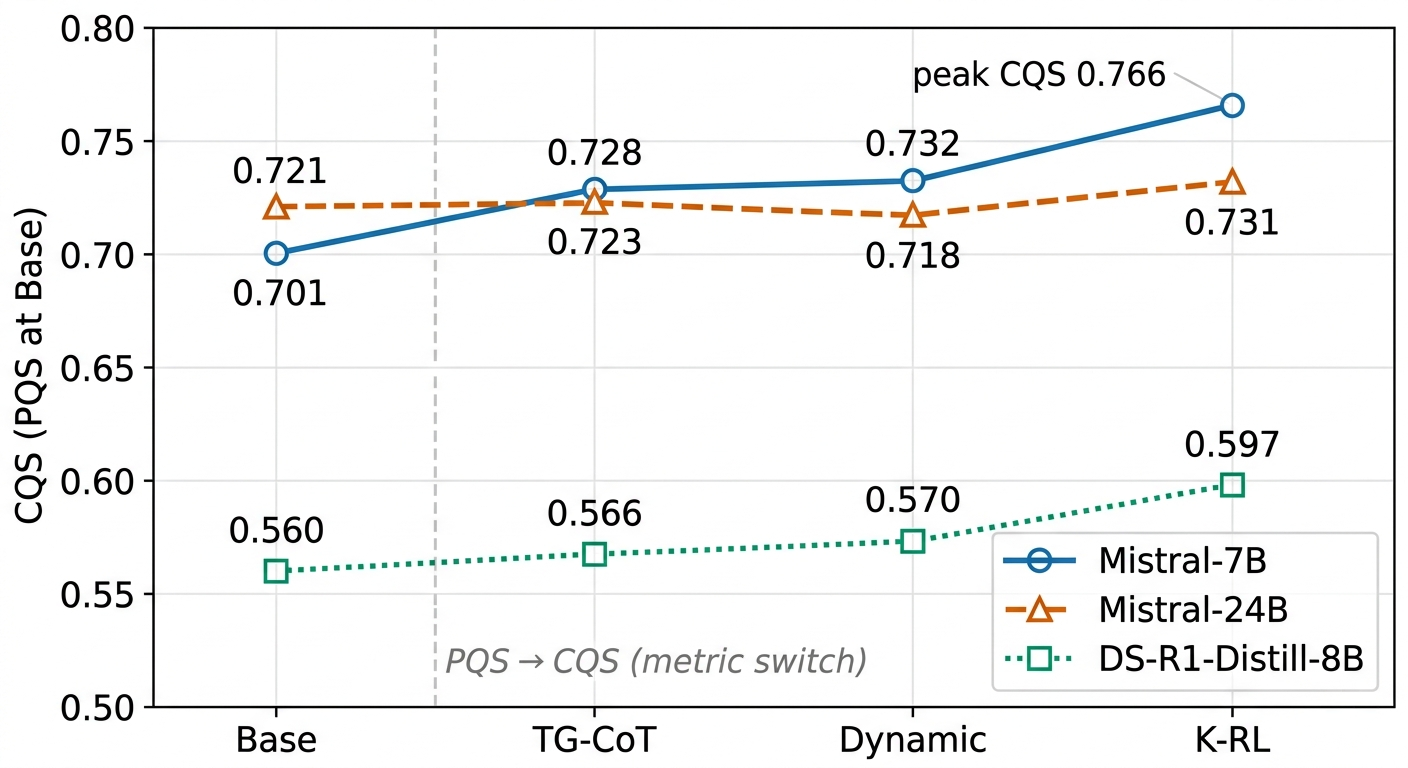}
    \caption{Stage-wise CQS trajectories on representative medium-scale models across training stages.}
    \label{fig:cqs_trajectory}
\end{figure}

Figure~\ref{fig:cqs_trajectory} and Table~\ref{tab:ablation} decompose the stage-wise trajectory on three representative models. PG-CoT (A$\to$B) introduces auditable structure with CQS reaching ${\sim}$0.72 on Mistral models; the apparent PQS drop is largely format bias. Dynamic SFT (B$\to$C) preserves CQS while adding longitudinal reasoning capability. K-RL (C$\to$D) further lifts CQS on all three models, with the largest gain on Mistral-7B. The progressive improvement confirms that each stage contributes incrementally to clinical quality (detailed scaling and model family analysis in Appendix~\ref{app:scaling}).

\paragraph{Format bias.}
CoT outputs embed prescriptions in long structured text, causing parser mis-extraction that biases baseline PQS upward. Detailed quantification is provided in Appendix~\ref{app:format_bias}.

\section{Conclusion}

We proposed a progressive four-stage framework (SFT $\to$ PG-CoT $\to$ Dynamic $\to$ K-RL) that targets three clinical gaps in LLM-based TCM prescription generation: the absence of auditable reasoning, the lack of longitudinal follow-up adjustment, and hidden contraindication violations. By constraining CoT distillation under the \emph{li-fa-fang-yao} paradigm, modeling patient trajectories with \emph{sui zheng jia jian} transition reasoning, and encoding deterministic pharmacological rules as rule-based DPO preference signals constructed from synthetic safety pairs, our framework substantially improves TCM prescription quality, reasoning auditability, and adversarial safety across diverse model families and scales, demonstrating that paradigm-constrained fine-tuning is an effective strategy for specialized clinical domains.

% Limitations is mandatory for ACL/EMNLP submissions and is not page-counted.
% Use \section* (unnumbered) per ACL template convention.
\section*{Limitations}
This work has several limitations. First, the TCMSafetyChecker covers only explicit herb incompatibility rules (\emph{Shi Ba Fan}, \emph{Shi Jiu Wei}) and toxicity classifications; it does not assess clinical indication-level safety (e.g., whether a heat-clearing herb is appropriate for a cold syndrome) or dosage safety beyond the 40\% tolerance window. Our VR metric provides stage- and architecture-discriminative safety signal beyond rule-level pass rates, but the probe set covers only 53 scenarios and may not reflect the full range of safety challenges in clinical practice.

Second, the safety benefits of PG-CoT and K-RL remain model-dependent. K-RL achieves the lowest VR on some architectures (e.g., LLaMA-3.1-8B: 0.038; Mistral-7B: 0.271) but increases VR on Gemma-2-9B (0.132 $\to$ 0.189); PG-CoT sharply reduces VR on Gemma-2-9B (0.340 $\to$ 0.094) but can increase VR on others (e.g., Phi-3-Mini: 0.382 $\to$ 0.401). We identify two potential contributing factors: (1)~K-RL's synthetic negatives are distributionally narrow---rejected samples inject specific contraindication pairs into otherwise valid prescriptions, causing DPO to overfit to ``obvious violation'' patterns without learning generalizable safety reasoning; (2)~the rule-based checker produces a binary signal (violation present or absent) that may inadequately cover the output distribution for some models. For deterministic rule enforcement, a post-hoc safety checker remains the simpler and more reliable approach; K-RL's intended advantage of \emph{internalizing} safety preferences is only partially achieved.

Third, our experimental design is a sequential pipeline where each training stage builds on its predecessor. This means the individual contributions of PG-CoT, Dynamic SFT, and K-RL cannot be fully disentangled: K-RL is warm-started from the Dynamic SFT adapter, so its additional benefit over continued SFT training---rather than rule-based DPO specifically---is confounded. A controlled ablation (e.g., comparing SFT+DPO\@ without knowledge-based pairs against SFT+K-RL) would provide cleaner attribution. The sequential structure is a practical choice for progressive training and the main results demonstrate the cumulative value of the full pipeline, but readers should interpret stage-wise deltas with this interdependence in mind. Notably, the headline result of a 7B fine-tuned model surpassing zero-shot GPT-5 is primarily driven by domain SFT (Stage~1); PG-CoT, Dynamic SFT, and K-RL provide incremental improvements in reasoning auditability, longitudinal structure, and safety alignment on top of this SFT foundation.

Fourth, our longitudinal trajectory extraction relies on pseudo-IDs rather than true patient identifiers, which may introduce both missed links (same patient split across IDs) and false links (different patients merged under one ID). The unified test set contains only initial-visit cases (871 static); Dynamic models are therefore evaluated on the same static cases as PG-CoT models, so CQS differences reflect model capability rather than test-case type, and longitudinal reasoning capability is tested indirectly through maintained prescription quality plus qualitative case studies (Appendix~\ref{app:cases}). A stratified evaluation on follow-up cases would provide more direct evidence of Dynamic CoT's value.

Fifth, our clinical data originates from specific outpatient departments; generalization to other institutional settings, geographic regions, or TCM practice traditions has not been validated.

Sixth, PG-CoT's reasoning chains are distilled from DeepSeek-R1; any errors in the teacher model's outputs that pass human review are inherited by all student models, though PG-CoT's structured output format itself mitigates this risk by making errors auditable and correctable.

Seventh, while our expert evaluation (Appendix~\ref{app:human_eval}) confirms progressive quality improvement across stages (Cohen's $\kappa$ = 0.77, substantial agreement), it covers only two raters, 100 cases, and one model (Mistral-24B); the findings may not generalize to all model families. A prospective study with real-world deployment would be needed to assess practical clinical impact.

\section*{Ethics Statement}
This work uses de-identified clinical records from TCM outpatient departments for research purposes. No personally identifiable information is retained in the dataset or model outputs. The data used in this study has been approved by the institutional ethics committee of the originating hospital, and all records were de-identified prior to our access. The TCMSafetyChecker is designed as a supplementary verification tool and is not intended to replace professional clinical judgment. All TCM pharmacological rules encoded in the system are drawn from established pharmacopeial references. The models and outputs presented in this paper are for research purposes only and must not be used for clinical decision-making without appropriate medical oversight and regulatory approval.

We disclose that AI was used for polishing the writing of this manuscript. All scientific content, experimental design, and analysis were conducted by the authors.

\section*{Data Availability}
Upon publication, we will release an anonymized subset of the clinical data that protects patient privacy, together with training code, evaluation scripts, model adapters, the TCMSafetyChecker rule set, and the adversarial probe set to facilitate replication of our safety evaluation.

% Acknowledgments must NOT appear in the review version (double-blind).
% Uncomment for the camera-ready submission only.
% \section*{Acknowledgments}
% \input{acknowledgments}

\bibliography{references}

@misc{bai2022constitutional,
  title={Constitutional {AI}: Harmlessness from {AI} Feedback},
  author={Bai, Yuntao and Kadavath, Saurav and Kundu, Sandipan and Askell, Amanda and Kernion, Jackson and Jones, Andy and Chen, Anna and Goldie, Anna and Mirhoseini, Azalia and McKinnon, Cameron and others},
  year={2022},
  eprint={2212.08073},
  archivePrefix={arXiv},
  primaryClass={cs.CL},
  url={https://arxiv.org/abs/2212.08073},
  doi={10.48550/arXiv.2212.08073}
}

@article{deepseek2024v3,
  title={DeepSeek-{V3} Technical Report},
  author={{DeepSeek-AI} and Liu, Aojun and Feng, Bingxue and Wang, Bingxuan and Tang, Bo and Chen, Chong and Cheng, Chong and Dai, Junqi and Deng, Zhiyong and others},
  journal={Computing Research Repository},
  volume={arXiv:2412.19437},
  year={2024},
  url={https://arxiv.org/abs/2412.19437},
  doi={10.48550/arXiv.2412.19437}
}

@article{deepseek2025r1,
  title={DeepSeek-{R1}: Incentivizing Reasoning Capability in {LLMs} via Reinforcement Learning},
  author={{DeepSeek-AI} and Guo, Damai and Yang, Dejian and Zhang, Haowei and Song, Junxiao and Zhang, Ruoyu and Xu, Runxin and Zhu, Qihao and Ma, Shirong and Wang, Peiyi and others},
  journal={Computing Research Repository},
  volume={arXiv:2501.12948},
  year={2025},
  url={https://arxiv.org/abs/2501.12948},
  doi={10.48550/arXiv.2501.12948}
}

@inproceedings{dettmers2024qlora,
  title={{QLoRA}: Efficient Finetuning of Quantized {LLMs}},
  author={Dettmers, Tim and Pagnoni, Artidoro and Holtzman, Ari and Zettlemoyer, Luke},
  booktitle={Advances in Neural Information Processing Systems},
  volume={36},
  pages={10088--10115},
  year={2023},
  url={https://proceedings.neurips.cc/paper_files/paper/2023/hash/1feb87871436031bdc0f2beaa62a049b-Abstract-Conference.html},
  doi={10.5555/3666122.3668195}
}

@article{guo2022fangzheng,
  title={Principles of Relativity Between Prescription and Syndrome in {Treatise on Febrile Diseases}},
  author={Guo, Yu'na and Liu, Chao and Lian, Wenjing and Wang, Jie},
  journal={Chinese Journal of Experimental Traditional Medical Formulae},
  volume={28},
  number={22},
  pages={189--195},
  year={2022},
  doi={10.13422/j.cnki.syfjx.20222291}
}

@inproceedings{ho2023large,
  title={Large Language Models Are Reasoning Teachers},
  author={Ho, Namgyu and Schmid, Laura and Yun, Se-Young},
  booktitle={Proceedings of the 61st Annual Meeting of the Association for Computational Linguistics (Volume 1: Long Papers)},
  pages={14852--14882},
  year={2023},
  doi={10.18653/v1/2023.acl-long.830}
}

@inproceedings{hsieh2023distilling,
  title={Distilling Step-by-Step! Outperforming Larger Language Models with Less Training Data and Smaller Model Sizes},
  author={Hsieh, Cheng-Yu and Li, Chun-Liang and Yeh, Chih-Kuan and Nakhost, Hootan and Fujii, Yasuhisa and Ratner, Alexander and Krishna, Ranjay and Lee, Chen-Yu and Pfister, Tomas},
  booktitle={Findings of the Association for Computational Linguistics: ACL 2023},
  pages={8003--8017},
  year={2023},
  doi={10.18653/v1/2023.findings-acl.507},
  url={https://aclanthology.org/2023.findings-acl.507/}
}

@inproceedings{hu2022lora,
  title={{LoRA}: Low-Rank Adaptation of Large Language Models},
  author={Hu, Edward J. and Shen, Yelong and Wallis, Phillip and Allen-Zhu, Zeyuan and Li, Yuanzhi and Wang, Shean and Wang, Lu and Chen, Weizhu},
  booktitle={Proceedings of the International Conference on Learning Representations},
  year={2022},
  url={https://openreview.net/forum?id=nZeVKeeFYf9},
  doi={10.48550/arXiv.2106.09685}
}

@article{jiang2012syndrome,
  title={Syndrome Differentiation in Modern Research of Traditional {Chinese} Medicine},
  author={Jiang, Miao and Lu, Cheng and Zhang, Chi and Yang, Jing and Tan, Yong and Lu, Aiping and Chan, Kelvin},
  journal={Journal of Ethnopharmacology},
  volume={140},
  number={3},
  pages={634--642},
  year={2012},
  doi={10.1016/j.jep.2012.01.033},
  pmid={22322251}
}

@inproceedings{lee2024rlaif,
  title={{RLAIF} vs. {RLHF}: Scaling Reinforcement Learning from Human Feedback with {AI} Feedback},
  author={Lee, Harrison and Phatale, Samrat and Mansoor, Hassan and Mesnard, Thomas and Ferret, Johan and Lu, Kellie Ren and Bishop, Colton and Hall, Ethan and Carbune, Victor and Rastogi, Abhinav and Prakash, Sushant},
  booktitle={Proceedings of the 41st International Conference on Machine Learning},
  series={Proceedings of Machine Learning Research},
  volume={235},
  pages={26874--26901},
  year={2024},
  publisher={PMLR},
  url={https://proceedings.mlr.press/v235/lee24t.html},
  doi={10.48550/arXiv.2309.00267}
}

@article{long2013incompatibility,
  title={Study on Incompatibility of Traditional {Chinese} Medicine: Evidence from Formula Network, Chemical Space, and Metabolism Room},
  author={Long, Wei and Zhang, Xiao-Dong and Wu, Hong-Ying and Jin, Jin and Yu, Guang-Yun and He, Xin and Wang, Hao and Shen, Xiu and Zhou, Ze-Wei and Liu, Pei-Xun and Fan, Sai-Jun},
  journal={Evidence-Based Complementary and Alternative Medicine},
  volume={2013},
  pages={352145},
  year={2013},
  doi={10.1155/2013/352145}
}

@misc{meta2025llama4,
  title={The {Llama 4} Herd: Architecture, Training, Evaluation, and Deployment Notes},
  author={{Meta AI} and Dubey, Abhimanyu and Jauhri, Abhinav and Pandey, Abhinav and Kadian, Abhishek and Al-Dahle, Ahmad and Letman, Aiesha and Mathur, Akhil and Schelhammer, Alan and others},
  year={2026},
  eprint={2601.11659},
  archivePrefix={arXiv},
  primaryClass={cs.CL},
  url={https://arxiv.org/abs/2601.11659},
  doi={10.48550/arXiv.2601.11659},
  note={Withdrawn from arXiv due to incorrect authorship}
}

@misc{openai2025gpt5,
  title={Introducing {GPT-5}},
  author={{OpenAI}},
  year={2025},
  howpublished={\url{https://openai.com/index/introducing-gpt-5/}}
}

@inproceedings{ouyang2022training,
  title={Training Language Models to Follow Instructions with Human Feedback},
  author={Ouyang, Long and Wu, Jeffrey and Jiang, Xu and Almeida, Diogo and Wainwright, Carroll and Mishkin, Pamela and Zhang, Chong and Agarwal, Sandhini and Slama, Katarina and Ray, Alex and others},
  booktitle={Advances in Neural Information Processing Systems},
  volume={35},
  pages={27730--27744},
  year={2022},
  url={https://proceedings.neurips.cc/paper_files/paper/2022/hash/c8758c03983620f1183864a3b8db36e9-Abstract-Conference.html},
  doi={10.5555/3666122.3668229}
}

@inproceedings{rafailov2023direct,
  title={Direct Preference Optimization: Your Language Model is Secretly a Reward Model},
  author={Rafailov, Rafael and Sharma, Archit and Mitchell, Eric and Manning, Christopher D. and Ermon, Stefano and Finn, Chelsea},
  booktitle={Advances in Neural Information Processing Systems},
  volume={36},
  pages={53728--53741},
  year={2023},
  url={https://proceedings.neurips.cc/paper_files/paper/2023/hash/a85b405ed65c6477a4fe8302b5e06ce7-Abstract-Conference.html},
  doi={10.5555/3666122.3668446}
}

@article{singhal2023large,
  title={Large Language Models Encode Clinical Knowledge},
  author={Singhal, Karan and Azizi, Shekoofeh and Tu, Tao and Mahdavi, S. Sara and Wei, Jason and Chung, Hyung Won and Castellon, Liam and Celi, Marzyeh and Dieng, Adji and Li, Rishi and Liu, Connie and Kumar, Anil and DeSalvo, Giorgia and Natarajan, Vivek and Karthikesalingam, Alan and Matias, Yossi and Webster, Dale and Corrado, Greg S. and Ng, Andrew Y.},
  journal={Nature},
  volume={620},
  number={7972},
  pages={172--180},
  year={2023},
  doi={10.1038/s41586-023-06291-2}
}

@article{singhal2025medpalm2,
  title={Toward expert-level medical question answering with large language models},
  author={Singhal, Karan and Tu, Tao and Gottweis, Juraj and Sayres, Rory and Wulczyn, Ellery and Amin, Mohamed and Hou, Le and Clark, Kevin and Pfohl, Stephen R. and Cole-Lewis, Heather and Liu, Connie and Eslami, Shashanka and Poon, Noah and Bhojwani, Mitesh and Newman, Chris and Shukla, Chris and Mousavi, Mahdi and Le, Phong and Wu, Yong and Zhang, Yun and Sulyman, Aisha and Bastiani, Lucas and Azer, Maya and Li, Tao and Green, Sarah and D'Amico, Michael and Chakrabarti, Anil and Semturs, Christopher and Corrado, Greg S. and Karthikesalingam, Alan and Natarajan, Vivek},
  journal={Nature Medicine},
  volume={31},
  number={3},
  pages={943--950},
  year={2025},
  doi={10.1038/s41591-024-03423-7}
}

@article{tu2024towards,
  title={Towards Generalist Biomedical {AI}},
  author={Tu, Tao and Azizi, Shekoofeh and Driess, Danny and Schaekermann, Mike and Amin, Mohamed and Chang, David and Chiu, Andrew and DeMichele, Gary and Feng, Xiao and Ghosh, Jon and Haldar, Aakanksha and Hassani, Isabelle and Kanj, Tony and Kapp, Khaled and Koyejo, Sanmi and Mahdavi, S. Sara and Matias, Yossi and McLean, Jim and Mukherjee, Subhrajit and Oh, Sami and Park, Kashyap and Patra, Vivek and Ryskina, Elena and Ryu, Hadrien and Schuler, Kara and Tu, Kavita and Yee, Joyce and Lee, Yong and Karthikesalingam, Alan and Natarajan, Vivek},
  journal={NEJM AI},
  volume={1},
  number={3},
  pages={AIoa2300138},
  year={2024},
  doi={10.1056/AIoa2300138}
}

@article{tu2025conversational,
  title={Towards conversational diagnostic artificial intelligence},
  author={Tu, Tao and Schaekermann, Mike and Palepu, Anil and Saab, Khaled and Freyberg, Jan and Tanno, Ryutaro and Wang, Amy and Li, Brenna and Amin, Mohamed and Cheng, Yong and Vedadi, Elahe and Tomasev, Nenad and Azizi, Shekoofeh and Singhal, Karan and Hou, Le and Webson, Albert and Kulkarni, Kavita and Mahdavi, S. Sara and Semturs, Christopher and Gottweis, Juraj and Barral, Joelle and Chou, Katherine and Corrado, Greg S. and Matias, Yossi and Karthikesalingam, Alan and Natarajan, Vivek},
  journal={Nature},
  volume={642},
  number={8067},
  pages={442--450},
  year={2025},
  doi={10.1038/s41586-025-08866-7},
  url={https://www.nature.com/articles/s41586-025-08866-7}
}

@article{wang2009fangzheng,
  title={Connotation and Principles of Prescription--Syndrome Correspondence},
  author={Wang, Jie and Xiong, Xingjiang},
  journal={Journal of Traditional Chinese Medicine},
  volume={50},
  number={3},
  pages={197},
  year={2009},
  note={Chinese: fangzheng duiying neihan ji yuanze tantao}
}

@misc{wang2023bentsao,
  title={{HuaTuo}: Tuning {LLaMA} Model with {Chinese} Medical Knowledge},
  author={Wang, Haochun and Liu, Chi and Xi, Nuwa and Qiang, Zewen and Zhao, Sendong and Qin, Bing and Liu, Ting},
  year={2023},
  eprint={2304.06975},
  archivePrefix={arXiv},
  primaryClass={cs.CL},
  url={https://arxiv.org/abs/2304.06975},
  doi={10.48550/arXiv.2304.06975},
  note={Model also known as {BenTsao} (bencao, materia medica)}
}

@inproceedings{wei2022chain,
  title={Chain-of-Thought Prompting Elicits Reasoning in Large Language Models},
  author={Wei, Jason and Wang, Xuezhi and Schuurmans, Dale and Bosma, Maarten and Ichter, Brian and Xia, Fei and Chi, Ed H. and Le, Quoc V. and Zhou, Denny},
  booktitle={Advances in Neural Information Processing Systems},
  volume={35},
  pages={24824--24837},
  year={2022},
  url={https://proceedings.neurips.cc/paper_files/paper/2022/hash/9d5609613524ecf4f15af0f7b31abca4-Abstract-Conference.html},
  doi={10.5555/3666122.3666174}
}

@book{xie2023formulas,
  title={Formulary Science},
  author={Xie, Ming},
  publisher={Science Press},
  address={Beijing, China},
  year={2023},
  isbn={9787030741561},
  note={Chinese: fangjixue (formulary science). National TCM higher-education textbook; discusses the unified li-fa-fang-yao framework}
}

@article{yang2023baichuan2,
  title={Baichuan 2: Open Large-scale Language Models},
  author={Yang, Aiyuan and Xiao, Bowen and Wang, Bingxuan and Wang, Bolin and Zhou, Bin and Li, Chang and Hao, Chao and Huang, Dehui and Lyu, Dong and Wei, Fan and others},
  journal={Computing Research Repository},
  volume={arXiv:2309.10305},
  year={2023},
  url={https://arxiv.org/abs/2309.10305},
  doi={10.48550/arXiv.2309.10305}
}

@article{yu2023corefangzheng,
  title={Research Strategy of Core Prescription--Syndrome Based on Disease--Syndrome--Treatment Integration},
  author={Yu, Yue and Jin, Zixuan and Luo, Fukun and Liu, Wei and Wang, Pengqian and Xiong, Xingjiang},
  journal={China Journal of Chinese Materia Medica},
  volume={48},
  number={10},
  pages={2626--2633},
  year={2023},
  doi={10.19540/j.cnki.cjcmm.20230203.502},
  note={Chinese: jiyu bing-zheng-zhi jiehe de hexin fangzheng yanjiu silu; explicitly discusses the li-fa-fang-yao clinical workflow}
}

@misc{yue2024tcmbench,
  title={{TCMBench}: A Comprehensive Benchmark for Evaluating Large Language Models in Traditional Chinese Medicine},
  author={Yue, Wenjing and Wang, Xiaoling and Zhu, Wei and Zhang, Xinyi and Wang, Yukun and Huang, Zhiheng and Li, Zihan and Li, Yixuan and Zhang, Teng and others},
  year={2024},
  eprint={2406.01126},
  archivePrefix={arXiv},
  primaryClass={cs.CL},
  url={https://arxiv.org/abs/2406.01126},
  doi={10.48550/arXiv.2406.01126},
  note={arXiv preprint}
}

@inproceedings{zeng2020meddialog,
  title={{MedDialog}: Large-scale Medical Dialogue Datasets},
  author={Zeng, Guangtao and Yang, Wenmian and Ju, Zeqian and Yang, Yue and Wang, Sicheng and Zhang, Ruisi and Zhou, Meng and Zeng, Jiaqi and Dong, Xiangyu and Zhang, Ruoyu and Fang, Hongchao and Zhu, Penghui and Chen, Shu and Xie, Pengtao},
  booktitle={Proceedings of the 2020 Conference on Empirical Methods in Natural Language Processing (EMNLP)},
  pages={9241--9250},
  year={2020},
  doi={10.18653/v1/2020.emnlp-main.743},
  url={https://aclanthology.org/2020.emnlp-main.743/}
}

@inproceedings{wang2023huatuogpt,
  title={{HuatuoGPT}, Towards Taming Language Model to Be a Doctor},
  author={Zhang, Hongbo and Chen, Junying and Jiang, Feng and Yu, Fei and Chen, Zhihong and Chen, Guiming and Li, Jianquan and Wu, Xiangbo and Zhiyi, Zhang and Xiao, Qingying and Wan, Xiang and Wang, Benyou and Li, Haizhou},
  booktitle={Findings of the Association for Computational Linguistics: EMNLP 2023},
  pages={10859--10885},
  year={2023},
  address={Singapore},
  publisher={Association for Computational Linguistics},
  doi={10.18653/v1/2023.findings-emnlp.725},
  url={https://aclanthology.org/2023.findings-emnlp.725/}
}

@article{zhang2022tcmie,
  title={Information Extraction from the Text Data on Traditional Chinese Medicine: A Review on Tasks, Challenges, and Methods from 2010 to 2021},
  author={Zhang, Tingting and Huang, Zonghai and Wang, Yaqiang and Wen, Chuanbiao and Peng, Yangzhi and Ye, Ying},
  journal={Evidence-Based Complementary and Alternative Medicine},
  volume={2022},
  pages={1679589},
  year={2022},
  doi={10.1155/2022/1679589},
  url={https://doi.org/10.1155/2022/1679589}
}

@article{zhang2025qwen3embedding,
  title={Qwen3 Embedding: Advancing Text Embedding and Reranking Through Foundation Models},
  author={Zhang, Yanzhao and Li, Mingxin and Long, Dingkun and Zhang, Xin and Qiao, Yanzhe and Zhang, Yun and Zhou, Hui and Chen, Jie and Chen, Junyang and others},
  journal={Computing Research Repository},
  volume={arXiv:2506.05176},
  year={2025},
  url={https://arxiv.org/abs/2506.05176},
  doi={10.48550/arXiv.2506.05176}
}

@book{zhang200shanghan,
  title={Treatise on Febrile Diseases and Miscellaneous Diseases ({Shang Han Lun})},
  author={Zhang, Zhongjing},
  year={ca. 200},
  note={Classic TCM text; foundational syndrome-differentiated prescribing}
}

\clearpage
\appendix
\section{Clinical Background and Gap Motivation}
\label{app:clinical_background}

This appendix provides the TCM domain knowledge underlying the three clinical gaps defined in Section~\ref{sec:intro}.

\subsection{The \emph{Li-Fa-Fang-Yao} Paradigm and the SR Gap}
\label{app:sr}
 
The \emph{li-fa-fang-yao} (\begin{CJK}{UTF8}{gbsn}理法方药\end{CJK}) framework names the four-stage clinical reasoning chain through which TCM moves from pathomechanism to medication: \emph{li} (pathomechanism established via syndrome differentiation), \emph{fa} (therapeutic principle), \emph{fang} (formula selection), and \emph{yao} (herbal medication and compatibility)~\citep{jiang2012syndrome}. National formulary curricula treat this as the unified logic linking syndrome differentiation to prescribing~\citep{xie2023formulas}, while prescription--syndrome (\emph{fang--zheng}) correspondence research formalizes the requirement that formulas align with pathogenesis rather than symptoms alone~\citep{wang2009fangzheng,guo2022fangzheng,yu2023corefangzheng}. Historical roots trace to Zhang Zhongjing's \emph{Treatise on Febrile Diseases} (\emph{Shang Han Lun}), which codified syndrome-differentiated treatment with relatively fixed formula--syndrome relations~\citep{zhang200shanghan,guo2022fangzheng}.

A TCM practitioner follows this structured diagnostic chain in clinical practice:
\begin{itemize}[nosep,leftmargin=*]
    \item \textbf{Syndrome differentiation} (\begin{CJK}{UTF8}{gbsn}辨证\end{CJK}): integrates symptoms, tongue coating, and pulse qualities to identify the pathogenic pattern (e.g., Liver Qi stagnation with Spleen deficiency).
    \item \textbf{Treatment principle} (\begin{CJK}{UTF8}{gbsn}立法\end{CJK}): establishes the therapeutic strategy (e.g., ``course the liver and fortify the spleen'').
    \item \textbf{Formula construction} (\begin{CJK}{UTF8}{gbsn}选方\end{CJK}): selects herbs under the \emph{jun-chen-zuo-shi} (\begin{CJK}{UTF8}{gbsn}君臣佐使\end{CJK}) hierarchy---king herb targeting the main syndrome, minister herbs assisting, assistant herbs treating secondary symptoms or reducing toxicity, and courier herbs guiding or harmonizing.
    \item \textbf{Safety verification} (\begin{CJK}{UTF8}{gbsn}配伍验证\end{CJK}): checks herb--herb incompatibilities, especially the absolute prohibitions in Appendix~\ref{app:sc}.
\end{itemize}

Current LLMs perform end-to-end symptom-to-prescription mapping, producing isolated conclusions without this reasoning chain. In clinical practice, a prescription without an auditable rationale is useless---a practitioner cannot verify its appropriateness, nor learn from or correct it. This is the \textbf{SR Gap}. The chain must be explicit, stepwise, and follow the fixed \emph{li-fa-fang-yao} paradigm rather than free-form text~\citep{wang2009fangzheng,yu2023corefangzheng}.

\subsection{Follow-Up Visits and the LA Gap}
\label{app:la}

Follow-up visits constitute a substantial share of TCM outpatient care. In a follow-up, the patient presents with a new set of symptoms that reflect the therapeutic response to the previous prescription. The clinician must:
\begin{enumerate}[nosep,leftmargin=*]
    \item Compare the current symptoms with those recorded at the previous visit.
    \item Identify which aspects have improved, which have persisted, and which new manifestations have appeared.
    \item Adjust the prescription accordingly---adding herbs to address remaining or new patterns, removing herbs that are no longer needed, and modifying dosages. This is \emph{sui zheng jia jian} (\begin{CJK}{UTF8}{gbsn}随证加减\end{CJK}, ``adding or subtracting according to the pattern'').
\end{enumerate}

Errors in this step are the most common source of medication problems in TCM. Existing LLM approaches treat each visit independently (static prescribing), completely missing the longitudinal dependency. A model that cannot reason about change from the previous prescription cannot safely support follow-up care. This is the \textbf{LA Gap}.

\subsection{Absolute Pharmacological Prohibitions and the SC Gap}
\label{app:sc}

TCM has two classic sets of herb--herb incompatibilities that are considered absolute~\citep{long2013incompatibility}:
\begin{itemize}[nosep,leftmargin=*]
    \item \textbf{Shi Ba Fan} (\begin{CJK}{UTF8}{gbsn}十八反\end{CJK}, Eighteen Incompatibilities): six core pairs with derived synonyms. The most clinically important are licorice (Gan Cao) $\leftrightarrow$ Gan Sui, Da Ji, Yuan Hua, Hai Zao; and aconite (Chuan Wu, Cao Wu, Fu Zi) $\leftrightarrow$ Bei Mu, Gua Lou, Ban Xia, Bai Ji.
    \item \textbf{Shi Jiu Wei} (\begin{CJK}{UTF8}{gbsn}十九畏\end{CJK}, Nineteen Mutual Antagonisms): ten pairs, e.g., Ding Xiang $\leftrightarrow$ Yu Jin; Ren Shen $\leftrightarrow$ Wu Ling Zhi; Rou Gui $\leftrightarrow$ Chi Shi Zhi.
\end{itemize}

These are pharmacological facts, not soft preferences. Combining such herbs can cause severe toxicity or complete loss of efficacy. LLMs that generate such pairs are clinically dangerous, yet often do so with the same confidence as safe prescriptions, and no built-in mechanism detects or blocks such outputs. However, standard RLHF treats safety as a preference to be learned from human labels, which is costly, inconsistent, and may still miss violations. A rule-based approach directly encoding these prohibitions would be more reliable---if it can be effectively integrated into alignment. This is the \textbf{SC Gap}.

\subsection{Why These Gaps Differ from General LLM Limitations}
\label{app:gap_distinct}

While general LLMs also suffer from hallucination and weak reasoning, the TCM context makes each gap clinically concrete:
\begin{itemize}[nosep,leftmargin=*]
    \item The \textbf{SR Gap} is not about any reasoning, but about reasoning that follows the \emph{li-fa-fang-yao} clinical paradigm~\citep{xie2023formulas,guo2022fangzheng}; free-form CoT is insufficient.
    \item The \textbf{LA Gap} is not about multi-turn dialogue (which LLMs handle well), but about comparing clinical states and justifying adjustments based on therapeutic feedback---causal reasoning over time.
    \item The \textbf{SC Gap} is not about general toxicity or harmful content, but about a small, closed set of deterministic pharmacological rules that general safety filters may miss because they are rare in general text.
\end{itemize}

These distinctions motivate the three specialized components (\tgcot{}, Dynamic, \krl{}) and the four-stage progressive training pipeline.

\FloatBarrier

\section{Training Hyperparameters}
\label{app:hyperparams}

Table~\ref{tab:hyperparams} summarizes the shared training configuration across all four stages.

\begin{table*}[tbp]
\centering
{\small
\begin{tabular}{ll}
\toprule
\textbf{Item} & \textbf{Setting} \\
\midrule
GPU & 1$\times$ NVIDIA A800-SXM4-80GB \\
Quantization & QLoRA 4-bit (NF4)~\citep{dettmers2024qlora} \\
Framework & Unsloth \\
LoRA targets & $q, k, v, o$-proj, gate, up, down-proj \\
LoRA $\alpha$ / dropout & 16 / 0 \\
\midrule
\multicolumn{2}{l}{\textit{SFT (Stages 1--3)}} \\
Epochs & 3 \\
Optimizer & AdamW 8-bit \\
LR scheduler & Linear \\
Max length & 2,048 \\
\midrule
\multicolumn{2}{l}{\textit{DPO (Stage 4)}} \\
Epochs & 1 \\
DPO $\beta$ & 0.1 \\
Learning rate & $5 \times 10^{-6}$ \\
Max prompt length & 1,024 \\
\midrule
\multicolumn{2}{l}{\textit{Scale-adaptive}} \\
LoRA $r$: $<$7B / 7--14B / $>$14B & 16 / 32 / 64 \\
LR: $<$7B / 7--14B / $>$14B & $2{\times}10^{-4}$ / $1{\times}10^{-4}$ / $5{\times}10^{-5}$ \\
Batch/GPU: $<$7B / 7--14B / $>$14B & 4 / 4 / 2 \\
Grad accum: $<$7B / 7--14B / $>$14B & 4 / 4 / 8 \\
\midrule
\multicolumn{2}{l}{\textit{Additional details}} \\
Weight decay & 0.01 (SFT only) \\
Warmup & 5 steps (SFT) / 10\% (DPO) \\
Precision & bf16 / fp16 \\
Gradient checkpointing & Unsloth Smart \\
Random seed & 3407 \\
\bottomrule
\end{tabular}
}
\caption{Training configuration. Larger models use higher LoRA rank and lower learning rates to prevent catastrophic forgetting, with unified effective batch size of 16. DPO is initialized from the Dynamic SFT adapter to preserve reasoning and longitudinal capabilities.}
\label{tab:hyperparams}
\end{table*}

\section{Data Pipeline Details}
\label{app:data_pipeline}

\begin{figure*}[tbp]
    \centering
    \includegraphics[width=\textwidth]{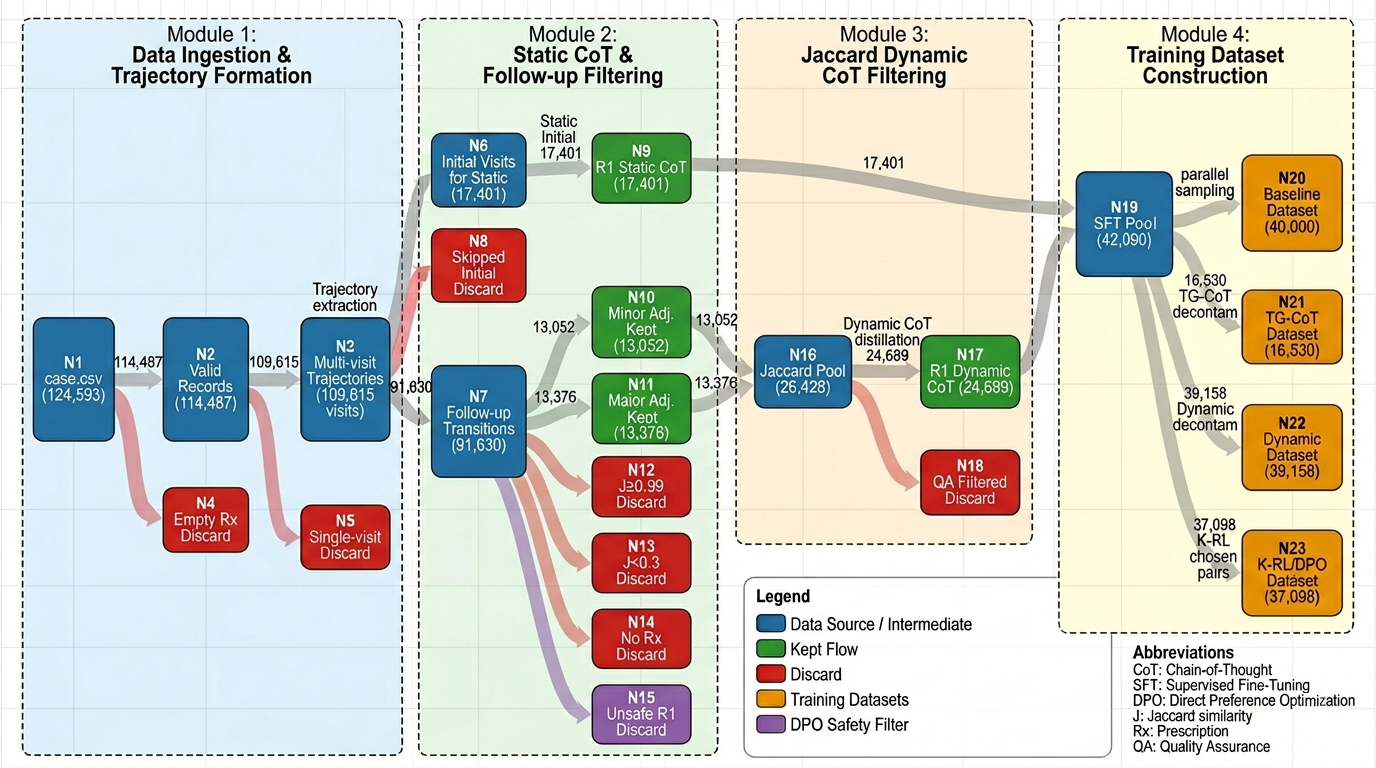}
    \caption{Data pipeline from 124,593 raw clinical records to the four training datasets. Trajectory extraction yields 17,985 patients (109,615 visits = 17,401 valid initial visits + 91,630 follow-up transitions + 584 skipped initial visits). Jaccard-based filtering retains 26,428 follow-up transitions; after R1 distillation and expert QA, 24,689 dynamic CoT samples join 17,401 static samples in a 42,090-sample SFT pool. Red streams denote discards; the four training datasets are parallel splits of this pool.}
    \label{fig:sankey}
\end{figure*}

Table~\ref{tab:dataset} and Figure~\ref{fig:sankey} summarize the data pipeline; dataset statistics are presented in Section~\ref{sec:data_models}.

\paragraph{Data source.}
The primary data source consists of 124,593 clinical encounter records from TCM outpatient departments, containing patient demographics, chief complaints, case histories, tongue and pulse descriptions, syndrome differentiations, and herbal prescriptions. Auxiliary resources include a TCM pharmacological knowledge base with standard herb names, aliases, properties (nature, flavor, channel entry), and toxicity classifications, along with a classical formula analysis compendium.

\paragraph{Entity alignment.}
Clinical prescriptions use diverse herb names and preparation forms (e.g., \begin{CJK}{UTF8}{gbsn}黑顺片\end{CJK} for Fu Zi, \begin{CJK}{UTF8}{gbsn}炒白术\end{CJK} for processed Bai Zhu). We normalize herb names through a three-tier matching strategy: exact match against standard names and aliases; prefix stripping of 18 common preparation forms (e.g., \begin{CJK}{UTF8}{gbsn}炒\end{CJK}/chao, \begin{CJK}{UTF8}{gbsn}炙\end{CJK}/zhi, \begin{CJK}{UTF8}{gbsn}制\end{CJK}/zhi, \begin{CJK}{UTF8}{gbsn}煅\end{CJK}/duan) followed by re-matching; and manual review of unmapped entries. This achieves 95.76\% coverage of clinical herb names, with 10,106 records lacking valid prescriptions excluded.

\paragraph{CoT distillation pipeline.}
We design the structured reasoning schema following the \emph{li-fa-fang-yao} paradigm and employ an LLM-auxiliary, expert-verified pipeline to construct training data. Specifically, we use DeepSeek-R1~\citep{deepseek2025r1} via the SiliconCloud Batch API (temperature $\tau{=}0.6$, $\texttt{max\_tokens}{=}4096$) to generate reasoning drafts in JSON format for each case record. These drafts then undergo extensive human review and correction: TCM practitioners verify syndrome differentiation accuracy, validate treatment principle--formula alignment, and correct pharmacological reasoning errors. This human-in-the-loop process ensures that the training data reflects clinically sound reasoning rather than unconstrained model generation. The pipeline yields 42{,}090 verified R1 reasoning samples across 9 batches (17{,}401 static + 24{,}689 dynamic) prior to train/test decontamination. The student model is then fine-tuned on these human-verified samples via supervised learning, learning to produce the structured reasoning chain before the prescription. This transforms the training objective from $P(\mathbf{y} \mid \mathbf{x})$ to $P(\mathbf{r}, \mathbf{y} \mid \mathbf{x})$, providing stronger gradient signal through intermediate supervision. Final post-decontamination training counts (16{,}530 PG-CoT and 39{,}158 Dynamic samples) are reported in Table~\ref{tab:dataset}.

\paragraph{Longitudinal extraction.}
Because the clinical dataset lacks unique patient identifiers, we construct pseudo-IDs using a two-tier strategy:
\begin{itemize}[nosep,leftmargin=*]
    \item Records with detailed case history ($>20$ characters): $\text{ID} = \text{gender} + \text{hash}(\text{history}[:100])$
    \item Records with brief history: $\text{ID} = \text{gender} + \text{inferred\_birth\_year}$
\end{itemize}
This yields 17,985 patient trajectories containing 109,615 total visits from 124,593 raw records. Single-visit patients (14,978) are excluded as they carry no follow-up signal.

We classify prescription changes between consecutive visits using Jaccard similarity of herb sets:
\begin{equation}
    J(\mathbf{y}^{(t)}, \mathbf{y}^{(t+1)}) = \frac{|\mathcal{H}^{(t)} \cap \mathcal{H}^{(t+1)}|}{|\mathcal{H}^{(t)} \cup \mathcal{H}^{(t+1)}|}
\end{equation}
where $\mathcal{H}^{(t)}$ denotes the herb set at visit $t$. We retain follow-ups with $0.3 \leq J < 0.9$:
\begin{itemize}[nosep,leftmargin=*]
    \item $J \geq 0.99$ (continuation): prescription nearly unchanged, carrying no adjustment signal.
    \item $0.7 \leq J < 0.9$ (minor adjustment): small additions/removals, the most common clinical scenario.
    \item $0.3 \leq J < 0.7$ (major adjustment): substantial formula change, indicating disease progression or therapeutic pivot.
    \item $J < 0.3$ (complete change): likely unrelated prescription, too noisy for training.
\end{itemize}
This filtering yields 24,689 dynamic follow-up samples from 91,630 total follow-up transitions.

\section{PG-CoT Pilot Experiment: Free-form vs.\ Paradigm-Guided CoT}
\label{app:pgcot_pilot}

Section~\ref{sec:tgcot} motivates PG-CoT by contrasting paradigm-guided reasoning against free-form CoT. Here we report the pilot experiment that quantifies this contrast.

\paragraph{Setup.}
We use Mistral-7B as the student model and the same DeepSeek-R1 teacher to distill two training sets of equal size from the same 500 held-out clinical cases. Training set~A (\textbf{Free-form CoT}) contains R1-generated reasoning drafts without structural constraints: the model is free to produce any narrative chain before the prescription. Training set~B (\textbf{PG-CoT}, ours) contains R1 drafts that are subsequently structured and verified against the \emph{li-fa-fang-yao} schema (syndrome analysis, treatment principle, formula rationale, safety verification), following the same expert-verified pipeline described in Section~\ref{sec:tgcot}. Both sets are then fine-tuned on Mistral-7B under identical hyperparameters (Table~\ref{tab:hyperparams}), and evaluated on the same unified test set of 871 cases.

\paragraph{Metrics.}
We compare the two variants on three dimensions: (1)~\textbf{PQS} for prescription quality; (2)~\textbf{Clinician-rated auditability}, the proportion of reasoning steps that a TCM practitioner can verify against the clinical record (rated by two licensed TCM practitioners on a random subset of 100 outputs); and (3)~\textbf{Logical inconsistency rate}, the proportion of outputs containing at least one reasoning step that contradicts a prior step (e.g., a cold-syndrome diagnosis followed by heat-clearing herbs).

\paragraph{Results.}
Table~\ref{tab:pgcot_pilot} presents the results. PG-CoT achieves substantially higher auditability (+0.29) and lower logical inconsistency ($-$0.12) than free-form CoT, while also improving prescription quality. This confirms that the \emph{li-fa-fang-yao} paradigm constraint serves as a performance-enhancing inductive bias rather than a limitation on model expressiveness.

\begin{table}[htbp]
\centering
{\small
\begin{tabular}{lccc}
\toprule
Variant & PQS $\uparrow$ & Auditability $\uparrow$ & Inconsistency $\downarrow$ \\
\midrule
Free-form CoT & 0.423 & 0.33 & 0.23 \\
PG-CoT (ours) & \textbf{0.651} & \textbf{0.62} & \textbf{0.11} \\
\bottomrule
\end{tabular}
}
\caption{Pilot experiment: Free-form CoT vs.\ PG-CoT on Mistral-7B. Auditability and Inconsistency are assessed via human comparison by two licensed TCM practitioners on a random subset of 100 outputs. Auditability = proportion of verifiable reasoning steps; Inconsistency = proportion of outputs with at least one self-contradictory step. Both variants use the same training data source and hyperparameters.}
\label{tab:pgcot_pilot}
\end{table}

\FloatBarrier

\section{Complete Model List}
\label{app:models}

Table~\ref{tab:app_models} lists all models tracked by the automated pipeline.

\begin{table*}[tbp]
\centering
{\small
\begin{tabular}{llll}
\toprule
Model Key & HuggingFace ID & Params & Family \\
\midrule
qwen3\_5\_0.8b & unsloth/Qwen3.5-0.8B & 0.8B & Qwen \\
qwen3\_5\_2b & unsloth/Qwen3.5-2B & 2B & Qwen \\
qwen3\_5\_9b & unsloth/Qwen3.5-9B & 9B & Qwen \\
qwen3\_5\_27b & unsloth/Qwen3.5-27B & 27B & Qwen \\
gemma2\_2b & unsloth/gemma-2-2b-it & 2B & Gemma \\
gemma2\_9b & unsloth/gemma-2-9b-it & 9B & Gemma \\
gemma2\_27b & unsloth/gemma-2-27b-it & 27B & Gemma \\
phi3\_mini & unsloth/Phi-3-mini-4k-instruct & 3.8B & Phi \\
llama3\_1\_8b & unsloth/Meta-Llama-3.1-8B-Instruct & 8B & LLaMA \\
mistral\_7b & unsloth/Mistral-7B-Instruct-v0.3 & 7B & Mistral \\
mistral\_24b & unsloth/Mistral-Small-24B-Instruct-2501 & 24B & Mistral \\
deepseek\_r1\_8b & unsloth/DeepSeek-R1-Distill-Llama-8B & 8B & LLaMA \\
baichuan2\_7b & baichuan-inc/Baichuan2-7B-Chat & 7B & Baichuan \\
\bottomrule
\end{tabular}
}
\caption{Models tracked by the automated pipeline. Baichuan2-7B is compared as a TCM-specific zero-shot baseline (Table~\ref{tab:sota_comparison}) rather than a fine-tuned model.}
\label{tab:app_models}
\end{table*}

\section{Metric Design Rationale}
\label{app:metric_rationale}

\paragraph{PQS component selection and weighting.}
PQS is co-designed with licensed TCM practitioners to reflect clinical priorities in prescription evaluation. Dosage accuracy ($\text{Acc}_{\text{dose}}$, 35\%) receives the highest weight because TCM adheres to the principle that ``dosage determines efficacy'': the same herb at different dosages can produce opposing therapeutic effects (e.g., Da Huang at low dose stops diarrhea, at high dose promotes purgation). Herb-set F1 ($F1_H$, 30\%) captures whether the correct herbs are selected, the primary clinical concern after dosage. Herb-set Jaccard ($J$, 15\%) supplements F1 with set-level overlap, penalizing both omission and commission. Herb count reasonableness ($\text{OK}_{\text{cnt}}$, 10\%) flags structurally abnormal prescriptions (e.g., single-herb or 30-herb outputs), as TCM formulas typically contain 5--15 herbs. The safety score ($S$, 10\%) provides a hard constraint gate: prescriptions with contraindication violations receive $S < 1$, ensuring that safety violations directly reduce PQS even when other components are high. The relatively lower weight of $S$ reflects its role as a necessary precondition rather than a discriminative quality signal: on the standard test set, $S$ saturates to 1.0 across stages, and its discriminative value lies in the adversarial VR metric rather than in PQS itself.

\paragraph{Why CQS introduces $A_{\text{reason}}$.}
PQS measures whether the prescription is correct, but cannot assess whether the model arrived at that prescription through clinically valid reasoning. This distinction is central to our SR Gap argument: a model that produces the right prescription via spurious correlations is clinically unsafe, because such shortcuts fail on out-of-distribution cases. CQS addresses this by introducing the reasoning alignment score $A_{\text{reason}}$, which quantifies how closely the model's structured \emph{li-fa-fang-yao} reasoning chain matches the expert-verified reference chain. The 0.6/0.4 weighting between $\text{PQS}_{\text{base}}$ and $A_{\text{reason}}$ reflects the primacy of prescription quality as the clinically verifiable outcome: a correct prescription that cures the patient is the ultimate clinical goal, and a correct prescription with auditable reasoning further enables practitioner oversight. We use the format-stable baseline SFT PQS ($\text{PQS}_{\text{base}}$) rather than stage-specific PQS across all stages to avoid conflating format bias artifacts (Appendix~\ref{app:format_bias}) with reasoning quality changes.

\paragraph{$A_{\text{reason}}$ implementation.}
\begin{equation}
\begin{split}
    \text{CQS} =\;& 0.6 \cdot \text{PQS}_{\text{base}} + 0.4 \cdot A_{\text{reason}}, \\
    A_{\text{reason}} =\;& \sum_{k} w_k \cdot \cos(\mathbf{e}_{k}^{\text{gt}}, \mathbf{e}_{k}^{\text{pred}})
\end{split}
\end{equation}
Section weights $w_k$: syndrome analysis / evolution (30\%), treatment principle / adjustment (20\%), formula rationale (35\%), safety check (15\%).
$A_{\text{reason}}$ is computed using Qwen3-Embedding-8B~\citep{zhang2025qwen3embedding}, loaded locally via \texttt{sentence-transformers}. For each CoT output, the four structured sections are extracted by JSON parsing and encoded; cosine similarity to the ground-truth section embeddings is weighted $(0.30, 0.20, 0.35, 0.15)$ and summed to yield $A_{\text{reason}}$. CQS is then $0.6\,\text{PQS}_{\text{base}} + 0.4\,A_{\text{reason}}$ on the same test instance. Outputs that fail JSON parsing receive $A_{\text{reason}} \approx 0$ on missing sections, which lowers CQS accordingly.

\paragraph{CQS interpretation caveat.}
CQS uses the format-stable baseline SFT PQS ($\text{PQS}_{\text{base}}$) as a fixed constant across all stages; cross-stage CQS improvements therefore reflect \emph{only} changes in $A_{\text{reason}}$, not in prescription quality. This design avoids conflating format bias artifacts with reasoning quality (Appendix~\ref{app:format_bias}), but it also means CQS is insensitive to genuine prescription quality degradation. For example, Mistral-7B PQS declines from 0.701 (baseline SFT) to 0.350 (K-RL) due to format bias, yet CQS rises from 0.728 to 0.766---the improvement is entirely attributable to the $A_{\text{reason}}$ term. Readers should consult stage-specific PQS values (Table~\ref{tab:full_results}) alongside CQS for a complete picture of model behavior. We further note that $A_{\text{reason}}$ measures semantic similarity to the R1-distilled reference reasoning, not clinical reasoning correctness per se; a model that independently discovers an equally valid but different diagnostic pathway would be penalized. The expert evaluation (Appendix~\ref{app:human_eval}) partially addresses this by calibrating automatic scores against human judgment.

\paragraph{Why VR is proposed.}
The rule-based safety score $S$ embedded in PQS serves as a necessary precondition for prescription validity, but on the standard held-out test set it saturates to 1.0 across all stages, providing no stage- or architecture-level discrimination on safety. This saturation occurs because standard test cases do not actively challenge model safety boundaries. VR is designed to probe the model's ability to \emph{maintain safety under adversarial pressure}: by constructing targeted prompts that explicitly or implicitly elicit contraindication violations, VR measures how robustly a model upholds pharmacological constraints when pressured to violate them. This adversarial framing provides the discriminative safety signal that $S$ cannot: VR ranges from 0.038 to 0.432 across our evaluation (Table~\ref{tab:adversarial}), revealing substantial stage- and architecture-dependent variation that would be invisible under the standard test paradigm. The three probe categories (direct requests, contextual triggers, open generation) are designed to test safety maintenance across increasing levels of subtlety, from explicit rule violation requests to scenarios where contraindications must be recognized without prompting.

\FloatBarrier

\section{Complete Results}
\label{app:full_results}

Table~\ref{tab:full_results} presents the complete evaluation across all 12 fine-tuned models and four training stages.

\paragraph{DS-R1-Distill-8B anomaly.}
DS-R1-Distill-8B exhibits an unusual PQS trajectory (0.560 $\to$ 0.268 $\to$ 0.649 $\to$ 0.605): the PG-CoT stage produces a sharp PQS drop followed by a Dynamic-stage surge above Baseline. This pattern is explained by its extremely low Structural Completeness (StrC = 0.09): the model's PG-CoT outputs fail to conform to the JSON schema, causing parser mis-extraction and severely deflating PQS (an extreme form of the format bias discussed in Appendix~\ref{app:format_bias}). Dynamic SFT partially resolves this formatting issue, enabling proper prescription extraction and restoring PQS above Baseline. The CQS trajectory (0.566 $\to$ 0.570 $\to$ 0.597) is far more stable, confirming that the PQS volatility is a parser artifact rather than a genuine quality shift.

\begin{table*}[tbp]
\centering
\setlength{\tabcolsep}{2.5pt}
{\footnotesize
\resizebox{\textwidth}{!}{%
\begin{tabular}{ll cccc cccc ccc cc}
\toprule
 & & \multicolumn{4}{c}{\textbf{PQS}} & \multicolumn{4}{c}{\textbf{Herb F1}} & \multicolumn{3}{c}{\textbf{CQS}} & \textbf{StrC} & \textbf{Safety} \\
\cmidrule(lr){3-6} \cmidrule(lr){7-10} \cmidrule(lr){11-13} \cmidrule(lr){14-15}
Model & Scale & Base & CoT & Dyn & KRL & Base & CoT & Dyn & KRL & CoT & Dyn & KRL & Dyn & All \\
\midrule
\multicolumn{15}{l}{\textit{Small models ($<7$B)}} \\
Qwen3.5-0.8B & 0.8B & 0.639 & 0.351 & 0.363 & 0.329 & 0.476 & 0.236 & 0.248 & 0.216 & 0.707 & 0.723 & 0.716 & 0.99 & 1.0 \\
Qwen3.5-2B & 2B & 0.699 & 0.390 & 0.400 & 0.363 & 0.531 & 0.268 & 0.277 & 0.291 & 0.711 & 0.715 & 0.702 & 1.0 & 1.0 \\
Gemma-2-2B & 2B & 0.604 & 0.395 & 0.405 & 0.420 & 0.454 & 0.266 & 0.264 & 0.294 & 0.387 & 0.376 & 0.553 & 0.32 & 1.0 \\
Phi-3-Mini & 3.8B & 0.643 & 0.401 & 0.367 & 0.380 & 0.406 & 0.306 & 0.291 & 0.314 & 0.422 & 0.442 & 0.501 & 0.16 & 1.0 \\
\midrule
\multicolumn{15}{l}{\textit{Medium models (7--14B)}} \\
Qwen3.5-9B & 9B & 0.449 & 0.321 & 0.294 & 0.213 & 0.195 & 0.186 & 0.195 & 0.145 & 0.584 & 0.522 & 0.519 & 0.00 & 1.0 \\
LLaMA-3.1-8B & 8B & 0.710 & 0.400 & 0.385 & 0.390 & 0.530 & 0.278 & 0.256 & 0.266 & 0.727 & 0.731 & 0.724 & 1.0 & 1.0 \\
Mistral-7B & 7B & 0.701 & 0.405 & 0.391 & 0.350 & 0.519 & 0.276 & 0.263 & 0.225 & 0.728 & 0.732 & 0.766 & 1.0 & 1.0 \\
DS-R1-Distill-8B & 8B & 0.560 & 0.268 & 0.649 & 0.605 & 0.290 & 0.189 & 0.583 & 0.515 & 0.566 & 0.570 & 0.597 & 0.09 & 1.0 \\
Gemma-2-9B & 9B & 0.717 & 0.377 & 0.382 & 0.365 & 0.547 & 0.253 & 0.260 & 0.240 & 0.721 & 0.728 & 0.729 & 1.0 & 1.0 \\
\midrule
\multicolumn{15}{l}{\textit{Large models ($>14$B)}} \\
Qwen3.5-27B & 27B & 0.566 & 0.310 & 0.375 & 0.361 & 0.503 & 0.322 & 0.319 & 0.388 & 0.572 & 0.577 & 0.581 & 0.06 & 1.0 \\
Gemma-2-27B & 27B & 0.683 & 0.388 & 0.391 & 0.379 & 0.510 & 0.260 & 0.258 & 0.248 & 0.676 & 0.683 & 0.701 & 0.07 & 1.0 \\
Mistral-24B & 24B & 0.721 & 0.380 & 0.375 & 0.393 & 0.550 & 0.257 & 0.255 & 0.277 & 0.723 & 0.718 & 0.731 & 1.0 & 1.0 \\
\bottomrule
\end{tabular}%
}}
\caption{Complete results across all evaluated models. StrC = Structural Completeness (for Dynamic stage); Safety score is 1.0 across all models and experiments.}
\label{tab:full_results}
\end{table*}

\section{Format Bias Quantification}
\label{app:format_bias}

Table~\ref{tab:format_bias} quantifies the format bias on PG-CoT outputs: dosage-pattern-only extraction from the \emph{fang-yi-jie-gou} section raises Herb F1 from ${\sim}0.26$ to ${\sim}0.73$ and reduces zero-F1 samples from ${\sim}400/871$ to ${\sim}10/871$, confirming that the apparent PQS decline associated with CoT is a parser artifact rather than a genuine quality drop.

The magnitude of format bias varies substantially across model families. All four models in Table~\ref{tab:format_bias} exhibit similar standard Herb F1 (${\sim}0.25$--$0.28$) and Rx-Only recovery (${\sim}0.73$--$0.76$), suggesting that the extraction error rate is relatively uniform among models that produce well-formed structured outputs. However, models with low Structural Completeness (StrC, Table~\ref{tab:full_results}) experience more severe bias: DS-R1-Distill-8B (StrC$=$0.09) shows a PQS swing of $0.560 \to 0.268$ at the PG-CoT stage, while Qwen3.5-27B (StrC$=$0.06) drops from $0.566 \to 0.310$. The CQS metric mitigates this by using the format-stable baseline PQS, ensuring that reasoning quality rather than parsing artifacts drives cross-stage comparisons.

\begin{table}[htbp]
\centering
\setlength{\tabcolsep}{3pt}
{\small
\begin{tabular}{lcccc}
\toprule
 & \multicolumn{2}{c}{\textbf{Herb F1}} & \textbf{Rx-Only} & \textbf{0-F1 Rec.} \\
\cmidrule(lr){2-3} \cmidrule(lr){4-4} \cmidrule(lr){5-5}
Model & Std & Rx-Only & PQS & (\#) \\
\midrule
Mistral-24B & 0.257 & 0.727 & 0.513 & 394 \\
LLaMA-3.1-8B & 0.278 & 0.746 & 0.523 & 350 \\
Mistral-7B & 0.276 & 0.758 & 0.528 & 357 \\
Gemma-2-9B & 0.253 & 0.740 & 0.520 & 385 \\
\bottomrule
\end{tabular}
}
\caption{Format bias quantification on PG-CoT outputs. ``Std'' = standard pipeline; ``Rx-Only'' = dosage-pattern-only extraction; ``0-F1 Rec.'' = number of zero-Herb-F1 samples recovered.}
\label{tab:format_bias}
\end{table}

\section{Small Model Structured Output Failures}
\label{app:small_models}

Some models fail to produce valid structured output parseable into the four required \emph{li-fa-fang-yao} sections (syndrome, principle, formula, safety). The failure mode is consistent: affected models begin the JSON structure correctly but produce verbose, repetitive text in the \emph{fang-yi-jie-gou} section (e.g., Gemma-2-2B repeating ``\begin{CJK}{UTF8}{gbsn}性味凉\end{CJK}'' (\textit{xing wei liang}, ``nature and flavor cool'') dozens of times), exceeding the generation length budget and leaving the output without a closing brace. This makes the output unparseable as JSON rather than failing to learn the reasoning structure---the syndrome and treatment principle sections are typically well-formed. Interestingly, scale alone does not predict success: Qwen3.5-0.8B and -2B achieve high $A_{\text{reason}}$ ($>$0.93) and CQS $\approx$0.71, DS-R1-Distill-8B reaches CQS $\approx$0.57--0.60 across CoT stages, while Gemma-2-27B remains low, indicating that format compliance depends on model architecture rather than parameter count. Qwen3.5-27B also exhibits low Structural Completeness (StrC = 0.06), suggesting that larger models in families with pre-existing format issues do not automatically resolve them.

\section{Case Studies}
\label{app:cases}

We present two representative cases from the test set, showing outputs from Mistral-24B across all four training stages.

\paragraph{Case 1: Initial visit with spleen qi deficiency and dampness.}
The patient presents with abdominal distension, loose stools, fatigue, pale tongue with white coating, and a weak pulse.

\textbf{Baseline}: The model outputs a prescription directly:
\begin{quote}
\small\texttt{\begin{CJK}{UTF8}{gbsn}党参15g 白术12g 茯苓15g 甘草6g 陈皮9g 半夏9g 砂仁6g\end{CJK}}\\
\quad\textit{Dang Shen 15g, Bai Zhu 12g, Fu Ling 15g, Gan Cao 6g, Chen Pi 9g, Ban Xia 9g, Sha Ren 6g}
\end{quote}
No diagnostic rationale is provided; the practitioner cannot verify why these herbs were selected.

\textbf{PG-CoT}: The model produces a structured reasoning chain followed by the prescription:
\begin{quote}
\small\texttt{[Syndrome] \begin{CJK}{UTF8}{gbsn}脾气虚兼湿---腹胀便溏，乏力，舌淡苔白，脉弱，符合脾气虚弱、运化失司之证。\end{CJK}}\\
\quad\textit{Spleen qi deficiency with dampness---abdominal distension, loose stools, fatigue, pale tongue with white coating, weak pulse, consistent with spleen qi weakness and impaired transportation.}\\
\texttt{[Principle] \begin{CJK}{UTF8}{gbsn}健脾益气，化湿和中。\end{CJK}}\\
\quad\textit{Fortify the spleen and replenish qi, transform dampness and harmonize the middle.}\\
\texttt{[Formula] \begin{CJK}{UTF8}{gbsn}以四君子汤健脾益气为君，二陈汤化湿为臣，砂仁醒脾为佐。\end{CJK}}\\
\quad\textit{Si Jun Zi Tang as king to fortify spleen and replenish qi; Er Chen Tang as minister to transform dampness; Sha Ren as assistant to revive the spleen.}\\
\texttt{[Safety] \begin{CJK}{UTF8}{gbsn}检查十八反、十九畏：无配伍禁忌。\end{CJK}}\\
\quad\textit{Checked Shi Ba Fan and Shi Jiu Wei: no contraindication combinations.}\\
\texttt{[Prescription] \begin{CJK}{UTF8}{gbsn}党参15g 白术12g 茯苓15g 甘草6g 陈皮9g 半夏9g 砂仁6g\end{CJK}}\\
\quad\textit{Dang Shen 15g, Bai Zhu 12g, Fu Ling 15g, Gan Cao 6g, Chen Pi 9g, Ban Xia 9g, Sha Ren 6g}
\end{quote}
The reasoning is auditable: each herb's role is justified within the \emph{li-fa-fang-yao} paradigm.

\textbf{Dynamic}: Same as PG-CoT for this initial visit (no prior prescription to adjust).

\textbf{K-RL}: Identical reasoning chain with safety verification more prominent; prescription unchanged.

\paragraph{Case 2: Follow-up visit with partial improvement after initial treatment.}
The patient received Si Jun Zi Tang for spleen qi deficiency. Two weeks later, abdominal distension has improved but loose stools persist; new symptom: night sweating.

\textbf{Baseline}: The model outputs a new prescription with no reference to the prior one:
\begin{quote}
\small\texttt{\begin{CJK}{UTF8}{gbsn}黄芪20g 白术12g 防风9g 浮小麦15g 牡蛎20g 甘草6g\end{CJK}}\\
\quad\textit{Huang Qi 20g, Bai Zhu 12g, Fang Feng 9g, Fu Xiao Mai 15g, Mu Li 20g, Gan Cao 6g}
\end{quote}

\textbf{Dynamic}: The model produces longitudinal reasoning:
\begin{quote}
\small\texttt{[Evolution] \begin{CJK}{UTF8}{gbsn}腹胀减轻提示脾气渐复，但便溏未愈、新增盗汗，说明脾虚及气虚卫外不固。\end{CJK}}\\
\quad\textit{Reduced distension indicates partial spleen qi recovery, but persisting loose stools and new night sweating suggest spleen deficiency and weak defensive qi.}\\
\texttt{[Adjustment] \begin{CJK}{UTF8}{gbsn}保留党参、白术、茯苓健脾基础；加黄芪、防风益气固表（玉屏风散意）；加浮小麦、牡蛎收敛止汗。去半夏、陈皮（湿已减，无需燥湿）。\end{CJK}}\\
\quad\textit{Retain Dang Shen, Bai Zhu, Fu Ling as spleen-fortifying base; add Huang Qi and Fang Feng to replenish qi and secure the exterior (Yu Ping Feng San intent); add Fu Xiao Mai and Mu Li to astringe and stop sweating. Remove Ban Xia and Chen Pi (dampness reduced, no further drying needed).}\\
\texttt{[New Formula] \begin{CJK}{UTF8}{gbsn}以四君子汤合玉屏风散加减，健脾固表止汗。\end{CJK}}\\
\quad\textit{Modified Si Jun Zi Tang combined with Yu Ping Feng San to fortify the spleen, secure the exterior, and stop sweating.}\\
\texttt{[Safety] \begin{CJK}{UTF8}{gbsn}检查十八反、十九畏：无配伍禁忌。\end{CJK}}\\
\quad\textit{Checked Shi Ba Fan and Shi Jiu Wei: no contraindication combinations.}\\
\texttt{[Prescription] \begin{CJK}{UTF8}{gbsn}党参15g 白术12g 茯苓15g 黄芪20g 防风9g 浮小麦15g 牡蛎20g 甘草6g\end{CJK}}\\
\quad\textit{Dang Shen 15g, Bai Zhu 12g, Fu Ling 15g, Huang Qi 20g, Fang Feng 9g, Fu Xiao Mai 15g, Mu Li 20g, Gan Cao 6g}
\end{quote}
The model explicitly compares the current and prior states, justifying each herb addition and removal.

\textbf{K-RL}: Same dynamic reasoning with safety verification; prescription preserved as no contraindication violations are detected.

\section{Expert Evaluation Design}
\label{app:human_eval}

Automatic metrics measure surface-level agreement with reference outputs, but TCM prescription adequacy requires domain expert judgment. Two prescriptions may share no common herbs yet both be clinically reasonable (\emph{tong bing yi zhi}---different treatments for the same disease), a scenario where Herb F1 and Jaccard incorrectly penalize one prescription. Similarly, CQS (via $A_{\text{reason}}$) measures semantic similarity to the R1-distilled reference reasoning, not the \emph{correctness} of the reasoning itself. We therefore design a structured expert evaluation protocol to validate our automatic metrics and directly assess clinical quality.

\paragraph{Evaluation protocol.}
We randomly sample 100 cases from the unified test set (50 initial visits and 50 follow-up visits) and present outputs from all four training stages of the best-performing model (Mistral-24B by automatic metrics). Outputs are anonymized as System A/B/C/D with randomized ordering to prevent stage identification.

Two TCM professionals (licensed practitioners or graduate-level TCM researchers) independently score each output on five dimensions using a 1--5 Likert scale:

\begin{enumerate}[nosep,leftmargin=*]
    \item \textbf{Syndrome accuracy (D)}: Does the syndrome differentiation match the patient's symptoms, tongue, and pulse? \emph{(CoT/Dynamic/K-RL only)}
    \item \textbf{Li-fa-fang-yao coherence (C)}: Is the reasoning chain logically connected from syndrome to treatment principle to formula to herbs? \emph{(CoT/Dynamic/K-RL only)}
    \item \textbf{Prescription adequacy (P)}: Are the herb selection and dosage appropriate for the identified syndrome? \emph{(All stages)}
    \item \textbf{Safety (S)}: Does the prescription contain any contraindication violations or toxicity concerns? \emph{(All stages)}
    \item \textbf{Adjustment flexibility (F)}: Is the follow-up modification clinically justified, with reasonable explanation for additions/removals? \emph{(Dynamic/K-RL only)}
\end{enumerate}

Baseline outputs lack reasoning chains; dimensions D and C are marked N/A for baseline. Static cases have dimension F marked N/A. We report inter-rater agreement via Cohen's $\kappa$, between-system comparisons via Wilcoxon signed-rank tests with Bonferroni correction, and auto-human metric correlation via Spearman's $\rho$ between automatic scores (PQS, CQS) and corresponding human dimensions (P, C).

\begin{table}[htbp]
\centering
{\small
\begin{tabular}{lccccc}
\toprule
Stage & D $\uparrow$ & C $\uparrow$ & P $\uparrow$ & S $\uparrow$ & F $\uparrow$ \\
\midrule
Baseline & N/A & N/A & 4.7 & 3.9 & N/A \\
PG-CoT & 4.1 & 4.5 & 4.3 & 4.2 & N/A \\
Dynamic & 4.6 & 4.6 & 4.5 & 4.1 & 4.5 \\
K-RL & 4.5 & 4.3 & 4.6 & 4.8 & 4.5 \\
\midrule
Cohen's $\kappa$ & \multicolumn{5}{c}{0.77} \\
\bottomrule
\end{tabular}
}
\caption{Expert evaluation results (1--5 Likert scale, mean across two raters and 100 cases). D = Syndrome accuracy; C = Li-fa-fang-yao coherence; P = Prescription adequacy; S = Safety; F = Adjustment flexibility. N/A = not applicable for that stage.}
\label{tab:human_eval}
\end{table}

Inter-rater agreement is substantial (Cohen's $\kappa$ = 0.77). Several patterns emerge from Table~\ref{tab:human_eval}. First, structured reasoning stages consistently outperform Baseline on prescription adequacy (PG-CoT 4.3, Dynamic 4.5, K-RL 4.6 vs.\ Baseline 4.7); the slight Baseline advantage on P reflects that Baseline outputs are concise prescription lists without the narrative overhead that can occasionally dilute herb selection focus. Second, K-RL achieves the highest safety score (4.8), confirming that DPO-based alignment with pharmacological rules effectively internalizes safety preferences, consistent with the VR reductions observed in Table~\ref{tab:adversarial}. Third, Dynamic achieves the best syndrome accuracy (4.6) and coherence (4.6), suggesting that longitudinal reasoning training also sharpens diagnostic precision on initial visits. Fourth, Baseline's safety score (3.9) is the lowest across all stages, underscoring the SC Gap: models without explicit safety mechanisms are more prone to contraindication violations. The expert evaluation thus validates the automatic metrics and confirms that each training stage contributes incrementally to clinical quality.

\section{Reproducibility}
\label{app:reproducibility}

All experiments use a fixed random seed (3407) across all models and training stages. Training is performed on a single NVIDIA A800-SXM4-80GB GPU per job, with total compute estimated at approximately 700 A800-hours (52 training jobs averaging ${\sim}13$ hours each). We plan to release training code, evaluation scripts, and model adapters upon publication. Clinical data cannot be publicly released due to patient privacy regulations, but we provide the TCMSafetyChecker rule set and adversarial probe set to facilitate replication of safety evaluation.

\section{Prompt Templates}
\label{app:prompts}

\paragraph{Baseline instruction:}
\texttt{"Please generate a TCM herbal prescription based on the patient's symptoms."}

\paragraph{PG-CoT / Dynamic instruction:}
\texttt{"Please analyze the patient's condition following the li-fa-fang-yao paradigm (syndrome differentiation, treatment principle, formula rationale, safety check) and generate an appropriate TCM herbal prescription."}

\paragraph{Follow-up (Dynamic) instruction:}
\texttt{"This is a follow-up visit. Based on the initial prescription and the patient's current symptoms, analyze the condition evolution, explain the adjustment rationale, and generate a modified prescription."}

\section{Zero-shot vs.\ Fine-tuned}
\label{app:zeroshot}

\begin{table*}[tbp]
\centering
{\small
\begin{tabular}{lcccc}
\toprule
 & \multicolumn{2}{c}{PQS} & \multicolumn{2}{c}{Herb F1} \\
\cmidrule(lr){2-3} \cmidrule(lr){4-5}
Model & Zero-shot & Base SFT & Zero-shot & Base SFT \\
\midrule
\multicolumn{5}{l}{\textit{Small models ($<$7B)}} \\
Qwen3.5-0.8B & 0.344 & 0.639 & 0.132 & 0.476 \\
Qwen3.5-2B & 0.472 & 0.699 & 0.171 & 0.531 \\
Gemma-2-2B & 0.160 & 0.604 & 0.034 & 0.454 \\
Phi-3-Mini & 0.108 & 0.643 & 0.004 & 0.406 \\
\midrule
\multicolumn{5}{l}{\textit{Medium models (7--14B)}} \\
Qwen3.5-9B & 0.514 & 0.449 & 0.214 & 0.195 \\
LLaMA-3.1-8B & 0.200 & 0.710 & 0.051 & 0.530 \\
Mistral-7B & 0.113 & 0.701 & 0.002 & 0.519 \\
DS-R1-Distill-8B & 0.111 & 0.560 & 0.018 & 0.290 \\
Gemma-2-9B & 0.367 & 0.717 & 0.112 & 0.547 \\
\midrule
\multicolumn{5}{l}{\textit{Large models ($>$14B)}} \\
Mistral-24B & 0.396 & 0.721 & 0.147 & 0.550 \\
Gemma-2-27B & 0.353 & 0.683 & 0.125 & 0.510 \\
Qwen3.5-27B & 0.433 & 0.566 & 0.119 & 0.503 \\
\bottomrule
\end{tabular}
}
\caption{Zero-shot vs.\ baseline SFT on the unified test set (871 cases). Baichuan2-7B is reported in Table~\ref{tab:sota_comparison} as a TCM-specific zero-shot baseline.}
\label{tab:zeroshot}
\end{table*}

Zero-shot evaluation reveals substantial family-wise variation rather than a uniform ``near-zero'' capability. Chinese-centric models (Qwen3.5 at 0.8B--9B) reach 0.34--0.51 PQS without domain fine-tuning; Qwen3.5-27B reaches 0.433, while English-centric 7--8B models (Mistral, LLaMA, DeepSeek-R1-Distill) remain near 0.11--0.20. Baseline SFT still yields large gains on many English-centric models (e.g., Mistral-7B: 0.113 $\to$ 0.701 PQS) and moderate gains on Chinese-centric ones (e.g., Qwen3.5-27B: 0.433 $\to$ 0.566), though Qwen3.5-9B is an exception where zero-shot already exceeds baseline SFT on PQS; we attribute this to the model's strong Chinese-centric pretraining combining with a tendency to produce verbose, less structured outputs after SFT, which reduces dosage-accuracy extraction and thus PQS despite improved herb selection.

\section{Scaling and Model Family Analysis}
\label{app:scaling}

\paragraph{Scale effects.}
Among baseline models, larger models generally achieve higher PQS (Mistral-24B: 0.721 $>$ Mistral-7B: 0.701 $>$ Qwen3.5-0.8B: 0.639). This advantage diminishes in CoT experiments, where PQS alone is dominated by output format effects. CQS mitigates this: models with successful structured output achieve ${\sim}0.71$--0.73 CQS on PG-CoT despite PQS ${\sim}0.35$--0.40, because $A_{\text{reason}}$ remains high ($\sim$0.94--0.95). Models that fail JSON schema compliance (Gemma-2-27B) show low CQS regardless of scale; others can maintain moderate-to-high CQS despite depressed stage-wise PQS.

\paragraph{Model family effects.}
Mistral and LLaMA models show the most consistent performance, with Baseline PQS ${\sim}$0.70 and PG-CoT CQS ${\sim}$0.72--0.73. Qwen3.5-0.8B/2B match this pattern at ${\sim}$0.71--0.72 CQS; Qwen3.5-27B holds ${\sim}$0.57 CQS across CoT stages; Qwen3.5-9B holds ${\sim}$0.52 CQS on dynamic/k\_rl. Gemma-2-9B is stable (${\sim}$0.72--0.73 CQS); Gemma-2-2B PG-CoT CQS (${\sim}$0.39) remains below Baseline, while Gemma-2-27B stays low across stages.

\paragraph{Chinese-centric pretraining advantage?}
Contrary to intuition, Chinese-centric pretrained models (Qwen3.5 series) do not consistently outperform English-centric counterparts (Mistral, LLaMA) on this Chinese TCM task after domain SFT. This suggests that task-specific fine-tuning may matter more than pretraining language alone for final prescription quality.

\paragraph{Scale and safety.}
Adversarial VR remains substantial across scales (0.038--0.432 in Table~\ref{tab:adversarial}), with no architecture achieving near-zero violation rates under all stages. Mistral-7B shows the highest baseline VR (0.432); Gemma-2-9B shows the largest PG-CoT reduction (0.340 $\to$ 0.094). Stage rankings vary by family (K-RL lowest on 3/10 models; PG-CoT on several others), so safety conclusions should be reported per architecture rather than as a single global winner.

\end{document}